\RequirePackage{fix-cm}
\documentclass[twocolumn]{svjour3}          
\smartqed  
\usepackage{graphicx}
\usepackage[misc]{ifsym}
\usepackage{algorithm}
\usepackage{algpseudocode}
\usepackage{amsmath}
\usepackage{color}
\usepackage{multirow}
\usepackage{slashbox}
\usepackage{lineno,hyperref}
\usepackage{subfigure}
\usepackage{indentfirst}
\journalname{CGI2017} 
\begin{document}

\title{Histograms of Gaussian normal distribution for feature matching in clutter scenes}
\author{Wei Zhou \and Caiwen Ma \and Arjan Kuijper}
\institute{Wei Zhou (\Letter ) \and Caiwen Ma \at Xi'an Institute of Optics and Precision Mechanics of CAS, Xi'an, Shaanxi 710119, China\\ University of Chinese Academy of Sciences, No.19A Yuquan Road, Beijing 100049, China\\ email: zhouwei1@opt.cn \and Arjan Kuijper  \at Fraunhofer IGD, Fraunhoferstrae 5, 64283 Darmstadt, Germany\\ email: arjan.kuijper@igd.fraunhofer.de}
\date{ }

\maketitle

\begin{abstract}
3D feature descriptors 
provide  information between corresponding models and scenes.
3D objection recognition in cluttered scenes, however, remains a largely unsolved problem.
 Practical applications impose several challenges which are not fully addressed by existing methods.
Especially in cluttered scenes there are many feature mismatches between scenes and models.

We therefore propose Histograms of Gaussian Normal Distribution  (HGND) for extracting salient features on a local reference frame (LRF) that enables us to solve this problem.
We propose a LRF on each local surface patches
using the scatter matrix's eigenvectors. Then the HGND information of each salient point is calculated on the LRF, for which we use both the mesh and point data of the depth image.
Experiments on 45 cluttered scenes of the Bologna Dataset and 50 cluttered scenes of the UWA Dataset are made to evaluate the robustness and descriptiveness of our HGND.

Experiments carried out by us demonstrate that HGND obtains a more reliable matching rate than state-of-the-art approaches in cluttered situations.

\keywords{Local surface patch \and Local reference frame \and Local feature descriptor \and 3D feature matching }
\end{abstract}

\section{Introdution}
\label{sec:introdution}
 Among 3D data processing tasks, 3D object recognition has become one of the most popular researching problems in the last two decades \cite{zabulis2016correspondence,martinek2015Interachtive,Ahmed2015DTW,Guo2013Rotational,Tobari2010Unique,mian2010repeatability}. The main goals of object recognition are to correctly recognize objects in scenes and accurately estimate their poses \cite{mian2006three}. However, the depth information collected by the scanners most often contains noise, varying point density of point clouds, occlusions, and clutter. So recognizing  an object and recovering its pose from the recorded scenes are still a challenge in this research area.

Most of the recognition methods can be divided into several phases: feature points extraction, features calculation, feature matching, transform poses generation, and hypothesis verification \cite{taati2011local}. The key problem in 3D object recognition is how to describe the free-form object effectively, how to match these feature descriptors correctly,  how to recognize objects, and get their poses in the scenes. Therefore, the feature descriptor is the key to recognizing objects and its definition directly influences the subsequent phases of the recognition methods \cite{taati2011local}. 

\begin{figure}[htbp]
  \centering
  \includegraphics[width=8.5cm]{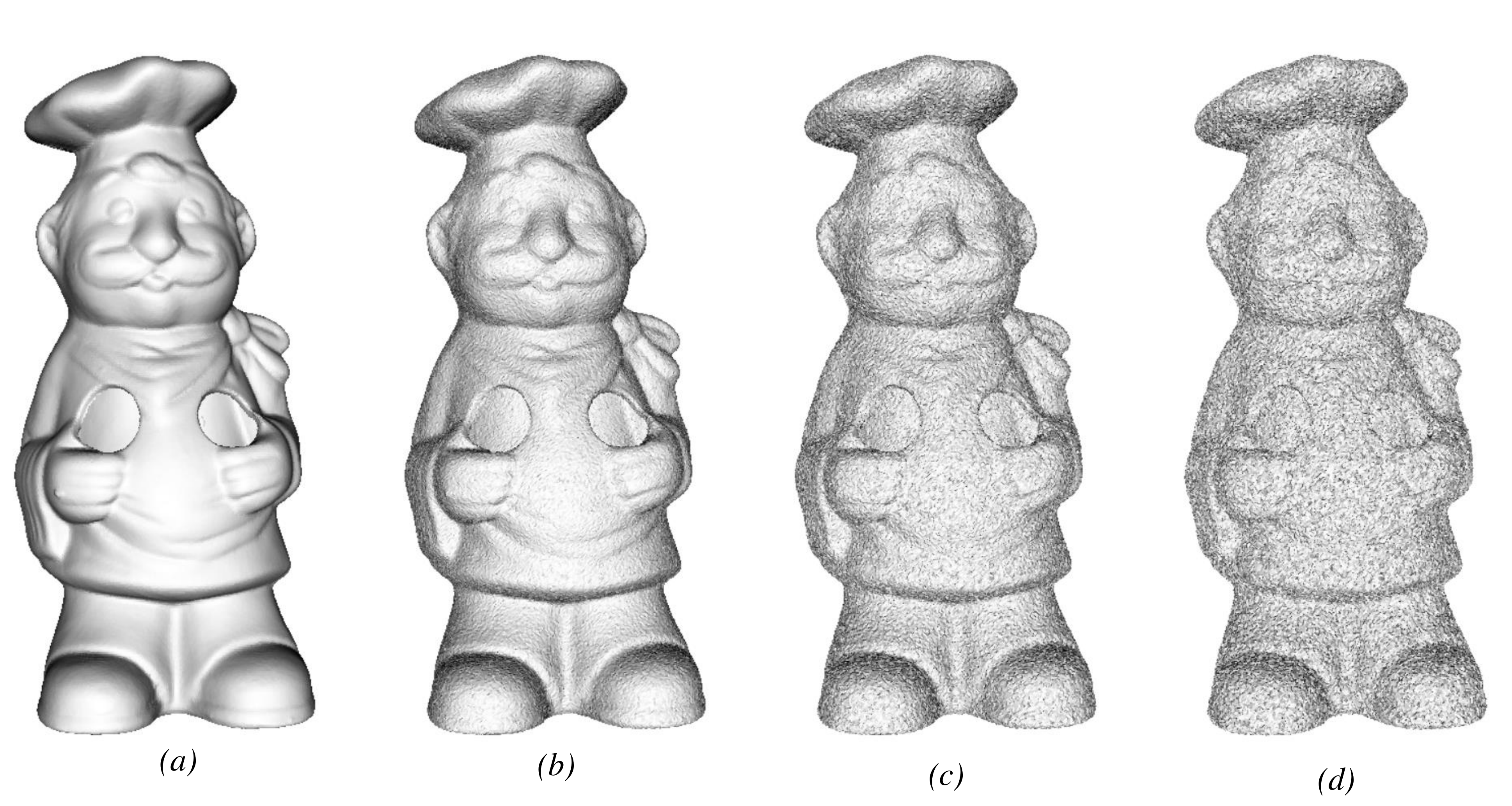}\\
  \caption{
	 \emph{(a)} Original Cheff model and with
     \emph{(b)} 0.1mr, 
     \emph{(c)} 0.2mr, and 
     \emph{(d)} 0.3mr Gaussian noise level. (UWA Dataset.)
	}\label{fig:gaussian_noise}
\end{figure}

\begin{figure}[htbp]
  \centering
  \includegraphics[width=8.5cm]{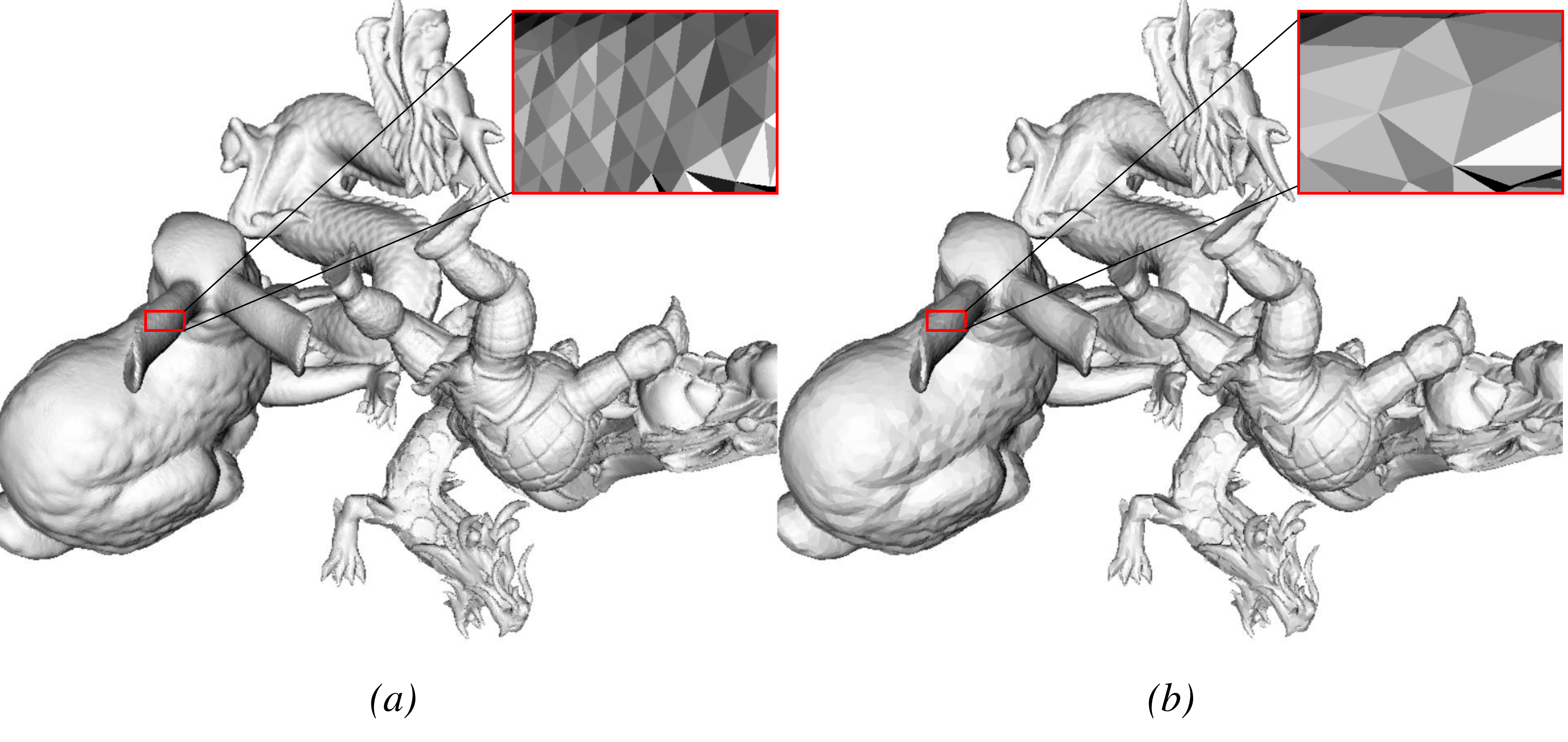}\\
  \caption{Cluttered scenes with different point density.
	\emph{(a)} Cluttered scenes with normal point density.
    \emph{(b)} Cluttered scenes with 1/8 of normal point density. (Bologna Dataset.)
	}\label{fig:point_density}
\end{figure}

The descriptiveness, robustness, and efficiency of the feature descriptors are the  three most important issues for feature matching \cite{bariya2010scale}. Due to the influence of the feature descriptor to the feature applications such as feature matching, transform generation, and object recognition, the  descriptiveness of the feature descriptor needs to be sufficiently high to ensure the accuracy of feature matching \cite{taati2011local}. Furthermore, the feature descriptor should be robust to the influence of a series of disturbances, such as noise (Sample demonstrating in \textbf{Fig.} \ref{fig:gaussian_noise}), varying point density, clutter, and occlusion (Sample seeing in \textbf{Fig.} \ref{fig:point_density}) \cite{Guo2013Rotational,boyer2011shrec}. In the remainder of the paper we use the wording ``cluttered scenes'' to describe scenes with such disturbances. In addition, the calculating efficiency of the feature descriptor should be high enough to decrease the calculation time of the algorithm.

The problem of feature matching in cluttered scenes is much harder than in normal non-interference scenes.
Significant limitations observed with state of the art methods are that their performances depend on whether the model is complete, e.g., occlusion and clutter exist in the scenes. Another difficulty is the point density of point clouds, as their feature matching method requires models and scenes under the same point density. In addition, existing literature focuses on evaluating the descriptors on noiseless data.

The motivation 
of our proposed technique is to 
convert the point data and mesh data information into a more descriptive and robust local feature representation that can decrease the feature mismatches between models and cluttered scenes.
If that has been done, the performances of many 
follow-up applications like 3D object recognition, 3D reconstruction, and 3D registration, will be improved.

We propose a novel technique to build  Local Reference Frames (LRFs) on 3D keypoints
in Section \ref{sec:local_reference_frame}
and present our so-called histograms of gaussian normal distribution (HGND) descriptor on the local surface patches in Section \ref{sec:Local_feature_descriptor}. A local surface patch is obtained by only considering the neighbor sphere surface around 3D keypoints from the range image. It thus consists of points and mesh data sets.
In Section \ref{sec:experiments} we show the effectiveness of the combination of LRF and HGND.

\section{Related work}
\label{sec:related_work}

According to the neighbor support radius, the existing feature descriptors can be divided into two main categories, global feature descriptors and local feature descriptors \cite{bariya2010scale,guo20143d,zhang2015pose}.
 The first category defines a series of features to describe the entire 3D object, whereas the latter one one use local parts of the object.

Global feature descriptors
ignore shape details and the object needs to be segmented from the scenes, they are not suitable for feature matching in cluttered scenes. On the other hand, the local feature description methods construct a series of features which describe the features of the local surface patches of feature points. So the local features are more robust to occlusion and clutter than global feature descriptor methods and it is suitable for feature matching in cluttered and occluded scenes \cite{petrelli2011repeatability}.

Several local-feature-based method have been proposed in the last decades, e.g.\cite{mian2006three}.
These methods can be divided into two categories by whether they construct a local reference frame (LRF) or not, before defining the feature descriptors \cite{Guo2013Rotational}.
Feature descriptors without LRF 
mostly adopt geometric information of the local surface to make up the feature. 

Transforming the geometrical information of local surface into a histogram by these methods that do not use a local reference frame causes most of the spatial information to be discarded.  This has direct negative consequences on the robustness and uniqueness of these methods. Therefore, local feature descriptors with LRF were proposed.
They are formed by the geometric information of feature points according to the local reference frame.

Tombari et al.\ \cite{Tobari2010Unique} introduced the \emph{signature of histograms of orientations} (SHOT) feature descriptor by computing local histograms incorporating geometric information of points. They proposed an LRF by calculating the eigenvector of the covariance matrix of the local neighboring surface of the feature points. By analyzing the importance of the LRF, they also proposed the weighted linear combination for calculating the covariance matrix and sign disambiguation. This method is invariant to rotation and translation, and robust to noise and clutter \cite{aldoma2012point}, but sensitive to varying point density \cite{Guo2013Rotational}.

Guo et al.\ \cite{Guo2013Rotational} introduced the \emph{rotational projection statistics} (ROPS) descriptor by rotationally projecting the neighboring points of the feature points into three tangent planes and calculating the statistics information of the projecting points. They also used the scatter matrix to form the LRF. 

Most of the proposed LRFs do not uniquely generate an invariant descriptor \cite{Tobari2010Unique} and they can thus not satisfy the requirements of descriptiveness, uniqueness, robustness and distinctiveness. So these may lead the descriptor to be sensitive to noise, varying point density, occlusion and clutter in the scene.
Inspired by these approaches, expecially ROPS \cite{Guo2013Rotational}  and  SHOT \cite{Tobari2010Unique} we combine the best parts and propose to construct an unambiguous LRF (Section \ref{sec:local_reference_frame}) and combine it with our well-performing statistic counting method, Histograms of Gaussian Normal Distribution (Section \ref{sec:Local_feature_descriptor}) in order to get higher recognition results in the feature matching applications (Section \ref{sec:experiments}).
Comparison is done with the two aforementioned LRF methods.


\section{Local reference frame}
\label{sec:local_reference_frame}

\begin{figure*}[htbp]
  \centering
  \includegraphics[width=17cm]{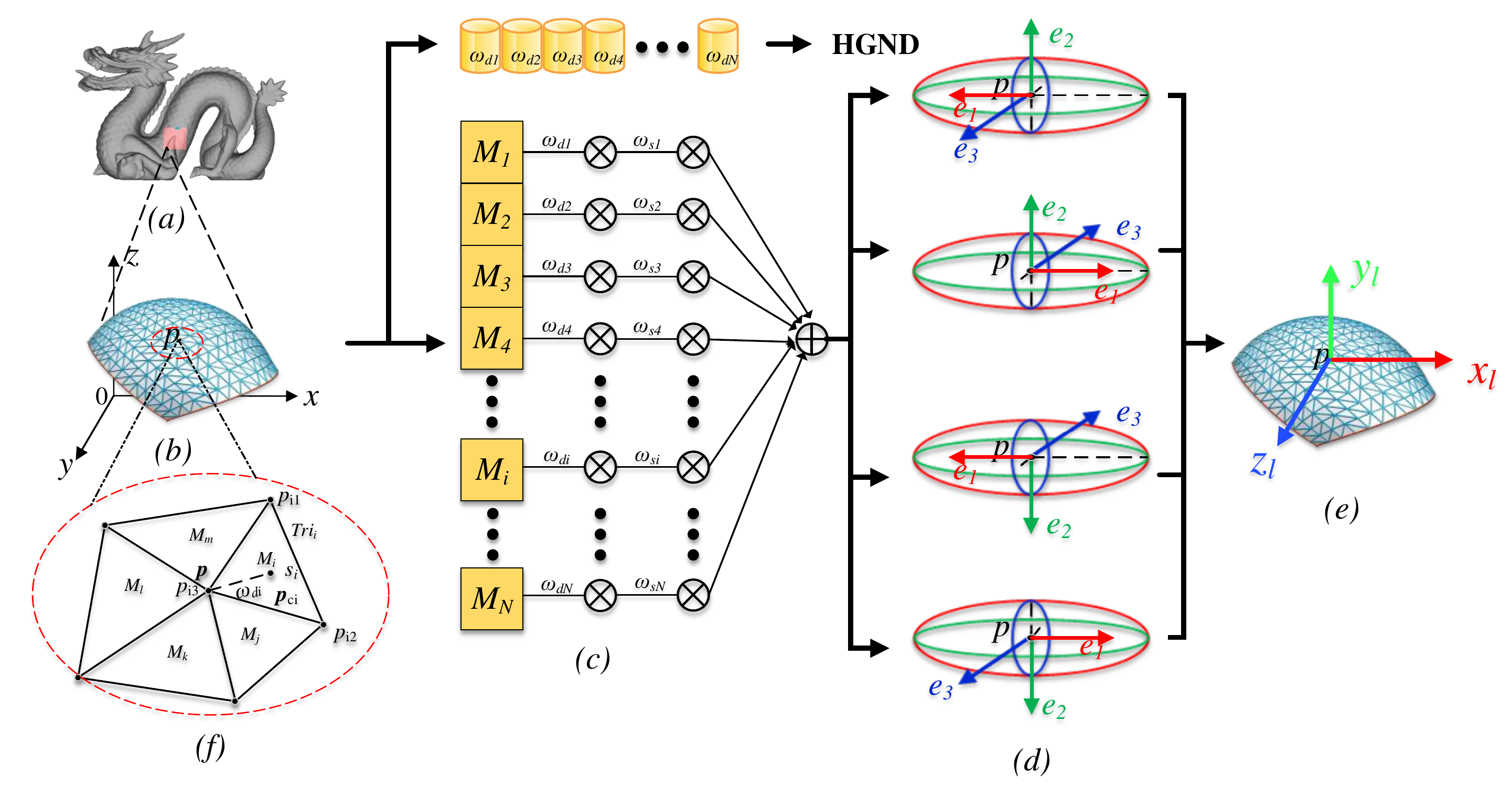}\\
  \caption{LRF generation processes.
	\emph{(a)} Asia Dragon of Bologna Dataset.
    \emph{(b)} The local surface patch is cropped from model.
    \emph{(c)} The scatter matrix $M$ of local surface patch is calculated by scatter matrix $M_{i}$ of each triangular mesh $Tri_{i}$.
    \emph{(d)} The most two descriptive eigenvectors $e_{1}, e_{3}$ are extracted from scatter matrix $M$.
    \emph{(e)} LRF is determined by sign disambiguation of $x$ and $z$ axes.
    \emph{(f)} Demonstrate triangular mesh of local surface patch.
	}\label{fig:LRF_PROCESS}
\end{figure*}

Before constructing the feature descriptor, we need to generate the local reference frame (LRF). 
In order to show the overall processes intuitively, the scheme of our LRF extraction is presented in Figure \ref{fig:LRF_PROCESS}.
Our LRF can roughly be divided into two parts: i) the calculation of scatter matrix \emph{\textbf{M}} and its most two descriptive eigenvectors (details in Section \ref{sec:scatter_matrix}); ii) the sign disambiguation of $x$ and $y$ axes (Section \ref{sec:sign_disambiguation}).
First, around a feature point $p$ on the depth image model or scene, a local surface patch is cropped. The scatter matrices are calculated for each triangle to get the scatter matrix of the local surface patch by distance and area weighted ($\omega_{di}, \omega_{si}$) summation. The $x$ and $y$ axes are extracted from scatter matrix \emph{\textbf{M}}, and we totally get 4 different LRFs. Then sign disambiguation is adopted both in $x$ axis and $y$ axis directions. Finally, the $z$ axis is obtained by the cross product of the $y$ and $x$ axes.

The distance weight $(\omega_{di})$ is also used as size weight to calculate HGND (Section \ref{sec:Local_feature_descriptor}).

\subsection{The calculation of scatter matrix \emph{\textbf{M}}}
\label{sec:scatter_matrix}

\begin{algorithm}
\caption{Calculation of the scatter matrix \emph{\textbf{M}}}\label{algorithm1}
\begin{algorithmic}[1]
\State \textbf{Input:} A local surface triangle mesh $m = (T, P)$, neighbor support radius $r$.
\State \textbf{Output:} Eigenvectors \{$\overrightarrow{\textbf{\emph{e}}_{1}}, \overrightarrow{\textbf{\emph{e}}_{2}}, \overrightarrow{\textbf{\emph{e}}_{3}}$\} of scatter matrix.
\Procedure {scatter matrix of lrf}{\emph{\textbf{M}}}.
\ForAll {$Tri_{i} \in m(T,P)$}
\State \textbf{Compute} the triangle centroid $p_{ci}$ and area $s_{i}$. 
\State \textbf{Compute} the distance weight $\omega_{di}$  and the area weight $\omega_{si}$, Using Eq. (\ref{eq:distance_weight}) and Eq. (\ref{eq:area_weight}).
\State \textbf{Compute} $M_{i}$ of each triangular mesh by integral transform Eq. (\ref{eq:final_scatter_matrix})
\State \textbf{Compute} the weighted summation \emph{\textbf{M}} by Eq. (\ref{eq:weightedM}).
\State \textbf{Decompose} $\emph{\textbf{M}}$ to get eigenvectors \{$\overrightarrow{\textbf{\emph{e}}_{1}}, \overrightarrow{\textbf{\emph{e}}_{2}}, \overrightarrow{\textbf{\emph{e}}_{3}}$\}.
\EndFor
\EndProcedure
\end{algorithmic}
\end{algorithm}

An outline of the calculation of the scatter matrix \emph{\textbf{M}} is given in \textbf{Algorithm} \ref{algorithm1} and \textbf{Fig.} \ref{fig:LRF_PROCESS}(c). Given a feature point $p$ and neighbor support radius $r$, the local surface triangle mesh of local surface patch is obtained  by cutting out the sphere surface of support radius $r$ and center $p$ from the range image. As is shown in Algorithm \ref{algorithm1} and \textbf{Fig.} \ref{fig:LRF_PROCESS}, our whole algorithm is calculated on the local surface triangle mesh to get the final local feature descriptor.

A random point of the triangle can be represented by
 $ p_{i}=ap_{i1}+bp_{i2}+cp_{i3}$ (see also \textbf{Fig.} \ref{fig:LRF_PROCESS}(f)),
where $a,b,c \in [0,1]$ and $a+b+c=1$.
So $p_{i}$ can also be expressed as
 $ p_{i}=ap_{i1}+bp_{i2}+(1-a-b)p_{i3}$.

For each triangle $i$ with vertices $p_{i1}, p_{i2}, p_{i3}$ we have the centroid $p_{ci}$  (see also \textbf{Fig.} \ref{fig:LRF_PROCESS}(f)) as:
$ p_{ci}=\frac{p_{i1}+p_{i2}+p_{i3}}{3}$.

The so-called scatter matrix \emph{\textbf{M}} is a statistical measure that is used to estimate the covariance matrix \cite{feller2008introduction}, represented by
$  \emph{\textbf{M}}=\sum_{i=1}^{N}(p_{i}-\overline{p})(p_{i}-\overline{p})^{T}$,
where $N$ is the number of  points in the local surface patch and $\overline{p}$ is the mean value of all these points.

As adaption of the definition of scatter matrix,
our scatter matrix $M$ of the local surface patch around the feature point $p$ is computed as follows:
\begin{equation}\label{eq:weightedM}
  \textbf{\emph{M}}=\frac{1}{\sum_{i=1}^{N}\omega_{si}}\cdot \frac{1}{\sum_{i=1}^{N}\omega_{di}}\cdot \sum_{i=1}^{N}\omega_{di}\omega_{si}M_{i},
\end{equation}
where $M_{i}$ is the scatter matrix of each triangle, $\omega_{di}, \omega_{si}$ are distance weight and area weight respectively. Different from the SHOT~\cite{Tobari2010Unique} and ROPS~\cite{Guo2013Rotational} methods, our distance weight of the triangle $\omega_{di}$ (see also \textbf{Fig.} \ref{fig:LRF_PROCESS}(f) and \textbf{Fig.} \ref{fig:gaussian_weight}(a)) is given by Gaussian function:
\begin{equation}\label{eq:distance_weight}
  \omega_{di}=\text{exp} \{- \frac{||p_{ci}-p||^2}{(2*\sigma_{d})^2} \},
\end{equation}
where $\sigma_{d}$ is the parameter of Gaussian function, in this paper we set $\sigma_{d}$ equal to 5mr.
The area $s_{i}$ of the triangle is given by  $
  s_{i}=|(p_{i2}-p_{i1})\times(p_{i3}-p_{i2})|/2$,
and the normalised area weight of each triangle then reads
\begin{equation}\label{eq:area_weight}
  \omega_{si}=\frac{s_{i}}{\sum_{i=1}^{N}s_{i}}{\color{red},}
\end{equation}
where the $\times$ denotes the cross product.

We now use definite integrals to calculate the scatter matrix of each triangle.
We so push all the points into the scatter matrix calculating process, and increase the calculation speed:
$  M_{i}= \int\int\int(p_{i}-\overline{p})(p_{i}-\overline{p})^{T}dxdydz / \\ \int\int\int dx\ dy\ dz
$. 
As we transformed the coordinate axes from $x,y,z$ coordinates to the $a,b$ coordinates, the triple integral is transformed into a double integral: 
\begin{equation}\label{16}
  M_{i}=\frac{\int_{0}^{1}\int_{0}^{1-a}(p_{i}(a,b)-\overline{p})(p_{i}(a,b)-\overline{p})^{T}dadb}{\int_{0}^{1}\int_{0}^{1-a}dadb}{\color{red},}
\end{equation}


In the computation process of the triangle's scatter matrix $M_{i}$,  we replace the mean point $\overline{p}$ with the local surface patch's feature point $p$, and substitute the triangle's points $p_{i}(a,b)$ with the triangle's vertexes $p_{i1},p_{i2},p_{i3}$ for increasing the calculation efficiency: 
\begin{equation}\label{eq:final_scatter_matrix}
\begin{split}
  M_{i}=\frac{1}{12}(\sum_{m=1}^{3}\sum_{n=1}^{3}(p_{im}-p)(p_{in}-p)^{T}+ \\
  \sum_{m=1}^{3}(p_{im}-p)(p_{im}-p)^{T}){\color{red},}
  \end{split}
\end{equation}
By applying Eq. (\ref{eq:weightedM}) on all $M_{i}$ we get \textbf{\emph{M}}. We apply an eigen decomposition 
on \textbf{\emph{M}} to get the eigenvalues and its corresponding eigenvectors \{$\overrightarrow{\textbf{\emph{e}}_{1}}, \overrightarrow{\textbf{\emph{e}}_{2}}, \overrightarrow{\textbf{\emph{e}}_{3}}$\}.

We select the largest two eigenvalues and its corresponding eigenvectors \{$\overrightarrow{\textbf{\emph{e}}_{1}}, \overrightarrow{\textbf{\emph{e}}_{2}}$\} to obtain the $x$ axis and $y$ axis. As showing in \textbf{Fig.} \ref{fig:LRF_PROCESS}(d), we totally get 4 different LRFs. In order to obtain feature descriptor's uniqueness, next section we'll present sign disambiguation of $x$ axis and $y$ axis.

\subsection{Sign disambiguation of LRF}
\label{sec:sign_disambiguation}


For the sign disambiguation of LRF, we only need to disambiguate the direction of $x$ and $y$ axes, and then using the cross product of $x$ and $y$ axes to get $z$ axis. Details are given in \textbf{Algorithm} \ref{algorithm2}.

\begin{algorithm}
\caption{Sign disambiguation of LRF}\label{algorithm2}
\begin{algorithmic}[1]
\State \textbf{Input:} Eigenvectors \{$\overrightarrow{\textbf{\emph{e}}_{1}}, \overrightarrow{\textbf{\emph{e}}_{2}}$\}, weight $\omega_{di}$ and $\omega_{si}$.
\State \textbf{Output:} LRF coordinate vector $(\emph{\textbf{x}}_{l}, \emph{\textbf{y}}_{l}, \emph{\textbf{z}}_{l})$.
\Procedure {sign disambiguation of } {$\emph{\textbf{x}}_{l},\emph{\textbf{y}}_{l}$}.
\ForAll {$Tri_{i} \in m(T,P)$}
\State \textbf{Compute} weighted product of $\overrightarrow{p_{ci}-p}\cdot\overrightarrow{\textbf{\emph{e}}_{2}}$.
\State \textbf{then} compute orientation function of $x$ and $y$ axes by Eq. (\ref{eq:x_direction}). 
\EndFor
\If {$ \text{sgn}(\emph{\textbf{x}}_{ori}) < 0 $}
\State $\textbf{\emph{x}}_{l} = \overrightarrow{\textbf{\emph{e}}_{1}}$
\Else
\State $\textbf{\emph{x}}_{l} = -\overrightarrow{\textbf{\emph{e}}_{1}}$
\EndIf
\If {$ \text{sgn}(\emph{\textbf{y}}_{ori}) < 0 $}
\State $\textbf{\emph{y}}_{l} = \overrightarrow{\textbf{\emph{e}}_{2}}$
\Else
\State $\textbf{\emph{y}}_{l} = -\overrightarrow{\textbf{\emph{e}}_{2}}$
\EndIf
\State \textbf{then} normalize $\overrightarrow{\textbf{\emph{e}}_{1}}, \overrightarrow{\textbf{\emph{e}}_{2}}$ (see Eq. (\ref{eq:xaxis})).
\State \textbf{Calculate} $\textbf{z}_{l}=\emph{\textbf{y}}_{l}\times\emph{\textbf{x}}_{l}$.
\EndProcedure
\end{algorithmic}
\end{algorithm}


%

We use the orientation function $\emph{\textbf{x}}_{ori}$ of $x$ axis to decide on the orientation of LRF's $x$ axis:
\begin{equation}\label{eq:x_direction}
  \emph{\textbf{x}}_{ori}=\frac{1}{\sum_{i=1}^{N}\omega_{si}}\cdot \frac{1}{\sum_{i=1}^{N}\omega_{di}}\cdot \sum_{i=1}^{N}\omega_{id}\omega_{is}\cdot\overrightarrow{p_{ci}-p}\cdot\overrightarrow{e_{1}},
\end{equation}
and
\begin{equation}\label{eq:xaxis}
  \emph{\textbf{x}}_{l}=\frac{\overrightarrow{\textbf{\emph{e}}_{1}}.\text{sgn}(\emph{\textbf{x}}_{ori})}{|\overrightarrow{\textbf{\emph{e}}_{1}}.\text{sgn}(\emph{\textbf{x}}_{ori})|},
\end{equation}

Similarly, the orientation function of $\emph{\textbf{y}}_{ori}$ of $y$ axis is defined by taking $\overrightarrow{\textbf{\emph{e}}_{2}}$ instead of $\overrightarrow{\textbf{\emph{e}}_{1}}$ in Eq.\ \ref{eq:x_direction}.
Finally, the $z$ axis is defined by cross product between $y$ and $x$ axes, (see also \textbf{Fig.} \ref{fig:LRF_PROCESS}(e)).

Our approach has several advantages.
For each triangle, the closer it is to the point $p$ and the larger the area is, the greater is the impact to the point $p$.
We use all triangle mesh points to calculate the LRF and due to the integral transformation and the integral computation of the triangles, the computational efficiency does not decrease.
Besides most of the existing methods don't determine the unique direction of LRF's axes. This leads to four LRFs and make the subsequent feature descriptor calculation process ambiguous.
They make the axis uniquely defined and result in the uniqueness and descriptiveness of the feature descriptor.

\section{Local Feature Descriptor}
\label{sec:Local_feature_descriptor}

\begin{figure*}[tbp]
  \centering
  \includegraphics[width=.9\hsize]{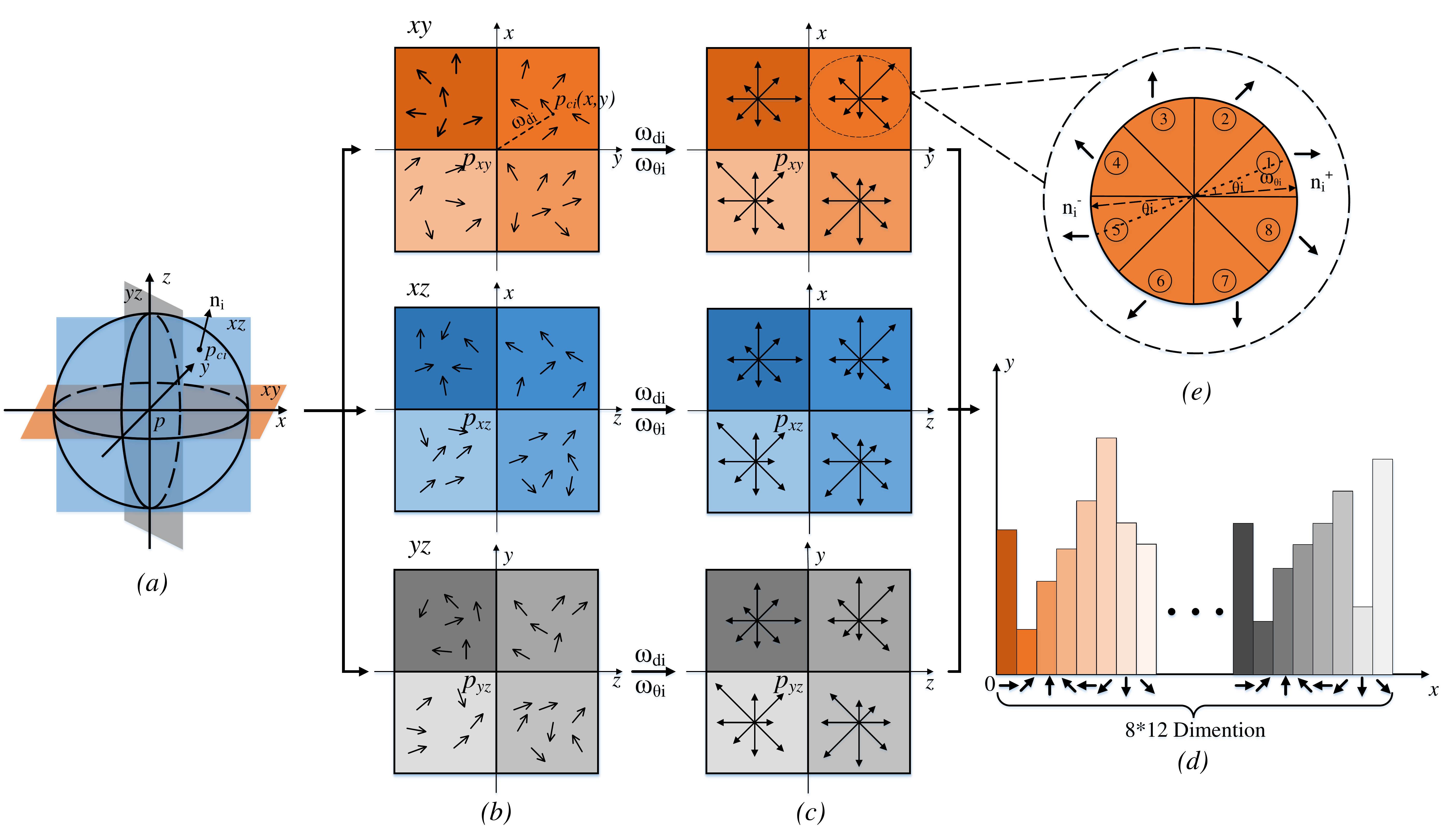}\\
  \caption{Feature descriptor generation processes.
    \emph{(a)} According to the LRF, the point data $\emph{\textbf{P}}(p_{1}, \ldots, p_{n})$ is transformed into $\emph{\textbf{P}}_{l} (p_{l1} ,\ldots,p_{ln})$; $\emph{\textbf{N}}(n_{1}, \ldots, n_{m})$ of each triangular mesh are obtained from the transformed data $\emph{\textbf{P}}_{l} (p_{l1} ,\ldots,p_{ln})$.
	\emph{(b)} The transformed data $\emph{\textbf{P}}_{l} (p_{l1} ,\ldots,p_{ln})$ is projected onto several planes to get the point projection data $\emph{\textbf{P}}_{prj}(\theta_{i}) (p_{prj1}, \ldots, p_{prjn})$.
	\emph{(c)} The area $\emph{\textbf{S}}(s_{1}, \ldots, s_{m})$ and normal $\emph{\textbf{N}}(n_{1}, \ldots, n_{m})$ of each triangular mesh are obtained from the LRF calculation processes.
	\emph{(d)} Both the geometrical information and the spatial distribution information are transferred into 2D histograms by linear interpolation.
	\emph{(e)} The 2D histograms are compressed and transferred into group information by Moment invariants and Shannon entropy.}\label{fig:HGND}
\end{figure*}

In the previous Section \ref{sec:local_reference_frame} we constructed the local reference frame for the local surface patch around the feature point $p$. In this section we will present the the subsequent stage: Generating the local feature descriptor in the local reference frame leading to our  Histograms of Gaussian Normal Distribution method.

In Section \ref{sec:related_work} we classified the descriptors into two categories: spatial information based and geometrical information based descriptors.
 We aim at a local feature descriptor which is descriptive, unique and robust to various kinds of occurring problems, viz.\ noise, clutter, occlusion and varying point density. We thus design our feature descriptor under these conditions and combine aspects of the two categories.
Clutter and occlusion mean that we consider scenes with multiple models that block each other.
Our descriptor is inspired by spatial descriptors, but such descriptor often perform weak on sparse data. In the computation process of the LRFs we have computed the Gaussian distance weight $\omega_{di}$ in the local surface patch. We will use this and Gaussian angle weight as the ``length" and ``direction" of transform normal distribution counting respectively (See also Figure \ref{fig:gaussian_weight}). in our descriptors compensating the defects of the spatial information based descriptor.

\textbf{Fig.} \ref{fig:HGND} shows the total generation processes of our feature descriptor coined ``Histograms of Gaussian Normal Distribution". From this figure, it's clear that our feature descriptor generation processes consists of two parts: data transform in 3D LRF (Section \ref{sec:data_transform}) and data counting in 2D surface (Section \ref{sec:data_counting}).

\subsection{Data transform in 3D}
\label{sec:data_transform}

Given a certain 3D object or scene, a local surface patch is cut around the feature points. The local surface patch includes the triangle mesh data $\emph{\textbf{T}}(t_{1}, \ldots, t_{m})$ and the point data $\emph{\textbf{P}}(p_{1}, \ldots, p_{n})$.
We then calculate the LRF $(\emph{\textbf{x}}_{l} ,\emph{\textbf{y}}_{l},\emph{\textbf{z}}_{l} )$ based on the local surface patch. According to the LRF, the point data $\emph{\textbf{P}}(p_{1}, \ldots, p_{n})$ is transformed to $\emph{\textbf{P}}_{l} (p_{l1}, \ldots, p_{ln})$, assuring rotation and translation invariance.
Finally the feature descriptors of each feature point on the LRF are calculated.

The transform point coordinate data is calculated by LRF matrix as
 $ p_{i1}^{'}=(\textbf{\emph{x}}_{l}, \textbf{\emph{y}}_{l}, \textbf{\emph{z}}_{l})\cdot(p_{i1}-p)$.
Similarly, we can get transformed coordinate data $p_{i2}^{'}, p_{i3}^{'}$ of other points  by LRF matrix. Then we obtain the transformed coordinate data $p_{ci}^{'}$ of the center point $p_{ci}$ by
 $   p_{ci}^{'}= p_{i1}^{'}+p_{i2}^{'}+p_{i3}^{'}/3$.

The normal $n_{i}^{'}$ of each transformed triangular mesh can be calculated by:
\begin{equation}\label{eq:trans_normal}
    n_{i}^{'}=\frac{(p_{i2}^{'}-p_{i1}^{'})\times(p_{i3}^{'}-p_{i2}^{'})}{|(p_{i2}^{'}-p_{i1}^{'})\times(p_{i3}^{'}-p_{i2}^{'})|},
\end{equation}

Based on Eq. (\ref{eq:trans_normal}) we can get the transformed normal data $\emph{\textbf{N}}^{'}(n_{1}^{'}, \ldots, n_{m}^{'})$ of local surface patch.

Then we project the $\emph{\textbf{N}}^{'}(n_{1}^{'}, \ldots, n_{m}^{'})$ and $\emph{\textbf{P}}_{c}^{'}(p_{c1}^{'}, \ldots, p_{cm}^{'})$ into the three coordinate planes $XY$, $XZ$ and $YZ$ of the LRF (See also \textbf{Fig.} \ref{fig:HGND}(b)), and obtain the projection data $\emph{\textbf{N}}_{xy}^{'}(n_{xy1}^{'}, \ldots, n_{xym}^{'})$,
$\emph{\textbf{N}}_{xz}^{'}(n_{xz1}^{'}, \ldots, n_{xzm}^{'})$, \\
$\emph{\textbf{N}}_{yz}^{'}(n_{yz1}^{'}, \ldots, n_{yzm}^{'})$ of
$\emph{\textbf{N}}^{'}$ as well as
$\emph{\textbf{P}}_{c_{xy}}^{'}(p_{c_{xy}1}^{'}, \ldots, p_{c_{xy}m}^{'})$,
$\emph{\textbf{P}}_{c_{xz}}^{'}(p_{c_{xz}1}^{'}, \ldots, p_{c_{xz}m}^{'})$,
$\emph{\textbf{P}}_{c_{yz}}^{'}(p_{c_{yz}1}^{'}, \ldots, p_{c_{yz}m}^{'})$ of
$\emph{\textbf{P}}_{c}^{'}$.

\subsection{Data counting in 2D projection surface}
\label{sec:data_counting}
We introduce an unique Gaussian weights group \{$\Omega_{di}, \Omega_{\theta i}$\} to count the normal histograms:
\begin{equation}\label{eq:gaussian_weights}
\begin{split}
\Omega_{di}=(\omega_{d_{xy}i}, \omega_{d_{xz}i}, \omega_{d_{yz}i}), \\
\Omega_{\theta i}=(\omega_{\theta_{xy}i}, \omega_{\theta_{xz}i}, \omega_{\theta_{yz}i}),
\end{split}
\end{equation}
The $(\omega_{d_{xy}i}, \omega_{\theta_{xy}i}), (\omega_{d_{xz}i}, \omega_{\theta_{xz}i}), (\omega_{d_{yz}i}, \omega_{\theta_{yz}i})$ are corresponding to three projection planes' ``length" and ``direction" weights respectively.

\begin{figure}[htbp]
  \centering
  \includegraphics[width=9cm]{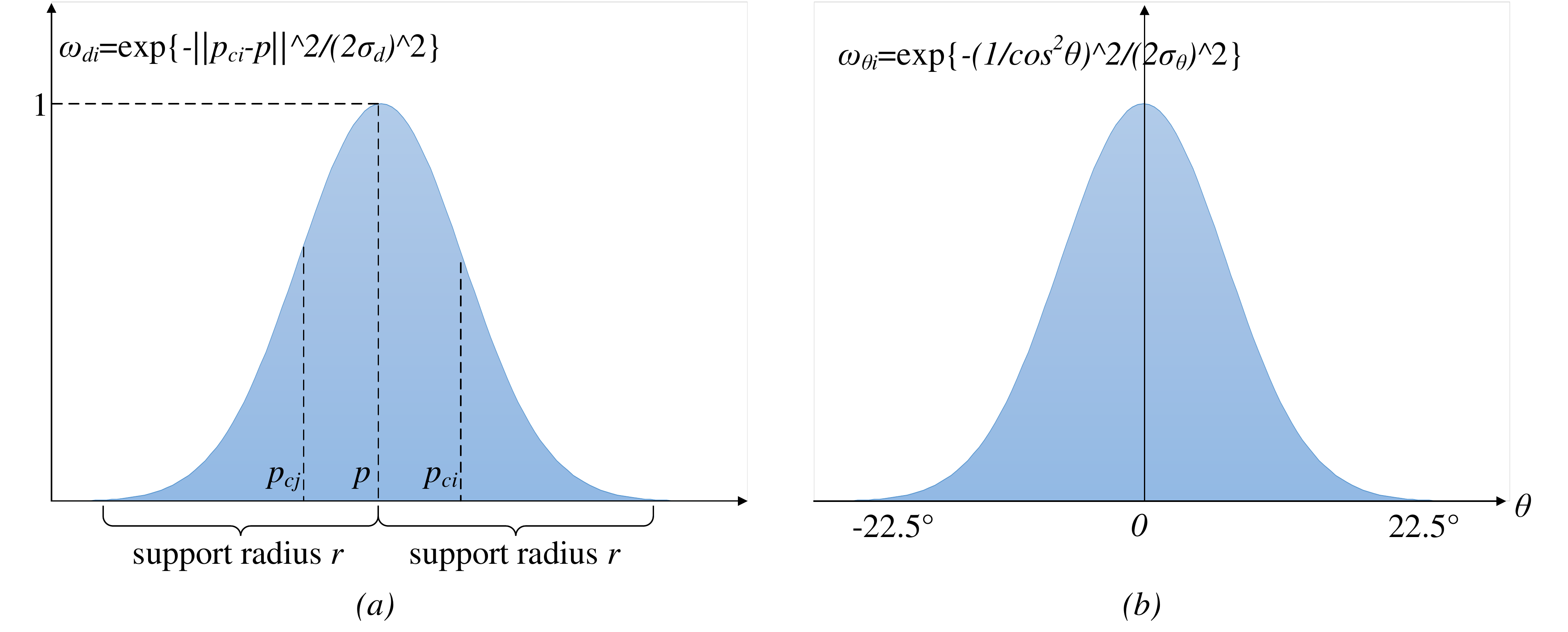}\\
  \caption{Gaussian weight function of ``length" and ``direction".
	}\label{fig:gaussian_weight}
\end{figure}

The calculation of ``length" Gaussian weight $\Omega_{di}$
is similar to $\omega_{di}$ (See also in Eq. (\ref{eq:distance_weight}) and \textbf{Fig.} \ref{fig:gaussian_weight}(a)):
\begin{equation}\label{eq:length_weight}
\Omega_{di}=\left\{
\begin{array}{lr}
  \omega_{d_{xy}i}=\text{exp} \{- ||p_{c_{xy}i}-p_{xy}||^2/(2*\sigma_{d})^2 \}, \\
  \omega_{d_{xz}i}=\text{exp} \{- ||p_{c_{xz}i}-p_{xz}||^2/(2*\sigma_{d})^2 \}, \\
  \omega_{d_{yz}i}=\text{exp} \{- ||p_{c_{yz}i}-p_{yz}||^2/(2*\sigma_{d})^2 \},
\end{array}
\right.
\end{equation}

From \textbf{Fig.} \ref{fig:gaussian_weight}(a) we can observe that for the center point $p_{ci}$ the more close to the feature point $p$ the more great weights $\omega_{di}$ it obtains, at the edge of local surface patch it obtains lowest weights. The projection plane's ``length" Gaussian weights $\Omega_{di}$
(Or 2D {``length"} Gaussian weight)
get similar to this case.

For the calculation of ``direction" Gaussian weight
$\Omega_{\theta i}$, we use
\begin{equation}\label{eq:original_direction_weight}
\omega_{\theta i}=\text{exp} \{- \frac{\theta_{i}^2}{(2*\sigma_{\theta})^2}\},
\end{equation}
where $\theta$ is the angle between normal and the center line of the $45^{\circ}$ sector, for calculation convenience we replace the numerator $\theta^2$ of Eq. (\ref{eq:original_direction_weight}) with $1/cos^{2}\theta$ (See also in \textbf{Fig.} \ref{fig:gaussian_weight}(b)):
\begin{equation}\label{eq:direction_weight}
\Omega_{\theta i}=\left\{
\begin{array}{lr}
  \omega_{\theta_{xy}i}=\text{exp} \{- 1/(2*cos\theta_{xy}i*\sigma_{\theta})^2\}, \\
  \omega_{\theta_{xz}i}=\text{exp} \{- 1/(2*cos\theta_{xz}i*\sigma_{\theta})^2\}, \\
  \omega_{\theta_{yz}i}=\text{exp} \{- 1/(2*cos\theta_{yz}i*\sigma_{\theta})^2\},
\end{array}
\right.
\end{equation}

From \textbf{Fig.} \ref{fig:gaussian_weight}(b) and \textbf{Fig.} \ref{fig:HGND}(e), we can observe that the angle $\theta$ between 2D normal $n_{i}$ and the center line of $45^{\circ}$ sector is range from $-22.5^{\circ}$ to $22.5^{\circ}$, the smaller the absolute value of the angle, the greater the weight the normal obtains. The projection plane's ``direction" Gaussian weight (Or 2D ``direction" Gaussian weight)
$\Omega_{\theta i}$ %
$(\omega_{\theta_{xy}i}, \omega_{\theta_{xz}i}, \omega_{\theta_{yz}i})$
get similar to this case.

For each of three projection planes, we calculate 2 level histograms.
As shown in \textbf{Fig.} \ref{fig:HGND}(b), firstly the point data $\emph{\textbf{P}}_{c}^{'}(p_{c1}^{'}, \ldots, p_{cm}^{'})$ is divided into 4 parts (4 quadrants) by its projection 2D coordinate value. At the same time, its corresponding ``length" Gaussian weights are also calculated.
For each quadrant, we divide it into 8 parts (8 direction) by the angles between the projection vectors of normal data $\emph{\textbf{N}}^{'}(n_{1}^{'}, \ldots, n_{m}^{'})$ and horizontal axis of 2D planes, and its corresponding ``direction" Gaussian weights are computed at the same time (See also in \textbf{Fig.} \ref{fig:HGND}(c), detail in \textbf{Fig.} \ref{fig:HGND}(e)). Specially, due to the uncertainty of normal direction, we also count once in each normal's opposite direction. For example, in \textbf{Fig.} \ref{fig:HGND}(e), the normal $n_{i}$ in the No.1 direction of the 8 parts we count it once in No.1 direction and also count it once in the opposite direction of No.5. One of the three projection planes $xy$'s calculation is presented in \textbf{Algorithm} \ref{algorithm3}, $xz,yz$ get similar to $xy$'s normal histogram calculation.

\begin{algorithm}
\caption{$XY$ calculation of Histograms of Gaussian Normal Distribution}\label{algorithm3}
\begin{algorithmic}[1]
\State \textbf{Input:} $\emph{\textbf{N}}^{'}(n_{1}^{'}, \ldots, n_{m}^{'})$, $\emph{\textbf{P}}_{c}^{'}(p_{c1}^{'}, \ldots, p_{cm}^{'})$.
\State \textbf{Output:} 4*8 dimension histograms.
\Procedure {$xy$ calculation of HGND} {4*8 dimension histograms}.
\For {$xy$ coordinate plane}
\ForAll {$Tri_{i} \in m(T,P)$}
\State \textbf{project} \{$n_{i}^{'}, p_{ci}^{'}$\} into $xy$ coordinate planes to get \{$n_{xyi}^{'}, p_{c_{xy}i}^{'}$\}.
\State \textbf{then} compute ($\omega_{d_{xy}i}, \omega_{\theta_{xy} i}$) by Eq. (\ref{eq:length_weight}) and Eq. (\ref{eq:direction_weight}).
\State \textbf{decide} the quadrant of $p_{c_{xy}i}^{'}$ in $xy$ plane.
\State \textbf{decide} the direction of $n_{xyi}^{'}$ in $p_{c_{xy}i}^{'}$'s corresponding quadrant.
\State \textbf{then} the value of \{$n_{xyi}^{'}, p_{c_{xy}i}^{'}$\}'s corresponding parts plus ($1*\omega_{d_{xy}i}*\omega_{\theta_{xy} i}$).
\State \textbf{then} the value of \{$n_{xyi}^{'}, p_{c_{xy}i}^{'}$\}'s opposite parts plus ($1*\omega_{d_{xy}i}*\omega_{\theta_{xy} i}$).
\EndFor
\EndFor
\State \textbf{store} the value of 4*8 parts from $xy$ projection data.
\EndProcedure
\end{algorithmic}
\end{algorithm}

Our approach has several advantages: 1) efficiency, only using mesh center point and mesh normal. Comparing to the normal methods, most of these methods calculate every point's normal by the neighbor points, these will result in large amount point calculation, like SHOT\cite{Tobari2010Unique} (See also 
\textbf{Fig.} \ref{fig:matching_time}). For one triangular mesh, ROPS\cite{Guo2013Rotational} uses total three points of every mesh, these will result in large calculation and also makes the feature descriptor sensitive to low point density (See also
 \textbf{Fig.} \ref{fig:point_density}). But we just use one center point of triangular mesh; 2) robustness, two Gaussian weights limit the influence of clutter. Eliminating the uncertainty of normal direction by ``double counting".

\section{Experiments}
\label{sec:experiments}


We use the 1-Precision (\emph{\textbf{FP}}/(\emph{\textbf{FP}} + \emph{\textbf{TP}})) and the Recall
(\emph{\textbf{TP}}/(\emph{\textbf{FN}} + \emph{\textbf{TP}})),
where $\emph{\textbf{FP}}$ ($\emph{\textbf{TP}}$) is the number of the False (True) Positives and $\emph{\textbf{FN}}$ is the number False Negatives.

{\color{red}For} fair comparison, we compute these values as follows:
	Given a model and a scene, we extract $1000$ points from the original model data and $n \times 1000$ points from the original scene data by uniform sampling, where $n$ is the number of  models in the scene.
	By extracting the corresponding points of the model keypoint from the scene keypoints, according to the given ground truth transformation (rotation and translation matrices), we retrieve these matches as $\emph{\textbf{TP+FN}}$, that is, all relevant matches.
	We calculate our feature descriptors for these keypoints and match the scene feature descriptors against all model feature descriptors. We find the nearest and second nearest model feature descriptors with a K-D tree \cite{bentley1975multidimensional}.
	If the ratio between the nearest distance and second nearest distance is less than a threshold $\epsilon$, the correspondence between the scene feature descriptor and model feature descriptor is marked as a $\emph{\textbf{TP+FP}}$, i.e.\ a selected element.
	We compare the selected elements, $\emph{\textbf{TP+FP}}$, with the corresponding matches $\emph{\textbf{TP+FN}}$ index to get the true positive ($\emph{\textbf{TP}}$) in the corresponding matches, and the other corresponding matches are signed as the false positive ($\emph{\textbf{FP}}$) and false negatives ($\emph{\textbf{FN}}$).
	The Recall vs 1-Precision Curve is obtained by varying the value of the corresponding match threshold.


Ideally the curves are located top-left, denoting high recall at low 1-precision. The curves can look complicated, though.

We use the Stanford 3D Scanning Repository Dataset \cite{levoy2005standford}, Bologna \cite{Tobari2010Unique,tombari2011combined,bolognadataset2010} (Sample demonstrating in \textbf{Fig.} \ref{fig:point_density}) and the UWA Datasets  (Seeing e.g. in \textbf{Fig.} \ref{fig:gaussian_noise}) \cite{mian2010repeatability,mian2006three,uwadataset2009} and compare our method against two state-of-the-art methods.
 All the methods are implemented in C++ and use the Point Cloud Library (PCL) \cite{rusu20113d}. PCL is a 3D point cloud processing software which includes the state-of-the-art methods and tools to deal with the 3D data and range image.



\subsection{Local Feature Descriptor Parameters}
\label{sec:local_descriptor_parameters}

There are three parameters in our feature descriptor calculation processes: support radius $\textbf{\emph{r}}$, length Gaussian weight $\sigma_{d}$ and direction Gaussian weight $\sigma_{\theta}$. The rotation angles and the projection surface affect the generation of the spatial feature information, while the 2D histogram bin number not only affects the geometrical feature information, but also determines the spatial feature information calculation.


According to the analysis for feature descriptor method, the \emph{support radius} determines the amount of feature information when describing the local surface in local feature descriptor: more large support radius implies more feature information obtained by descriptor, but this is only applicable to no clutter models and scenes. For the scenes with clutter, the support radius will have a critical value: less than this value, the more larger support radius implies more in formation obtained; greater than this value, the more larger support radius also implies more noise and other model's information are included.

We choose 6 support radii: 
(0.85, 4.25, 8.5, 17, 21.25, and 25.5mr), where ``mr" is mesh resolution, a common description in 3D mesh data processing denotes the mean length of the edges of the triangles of the 3D mesh \cite{petrelli2011repeatability}. We compare their effects by the Recall vs 1-Precision curves. In this experiment we keep the other parameters constant.

\begin{figure}[ht]
  \centering
  \includegraphics[width=4cm]{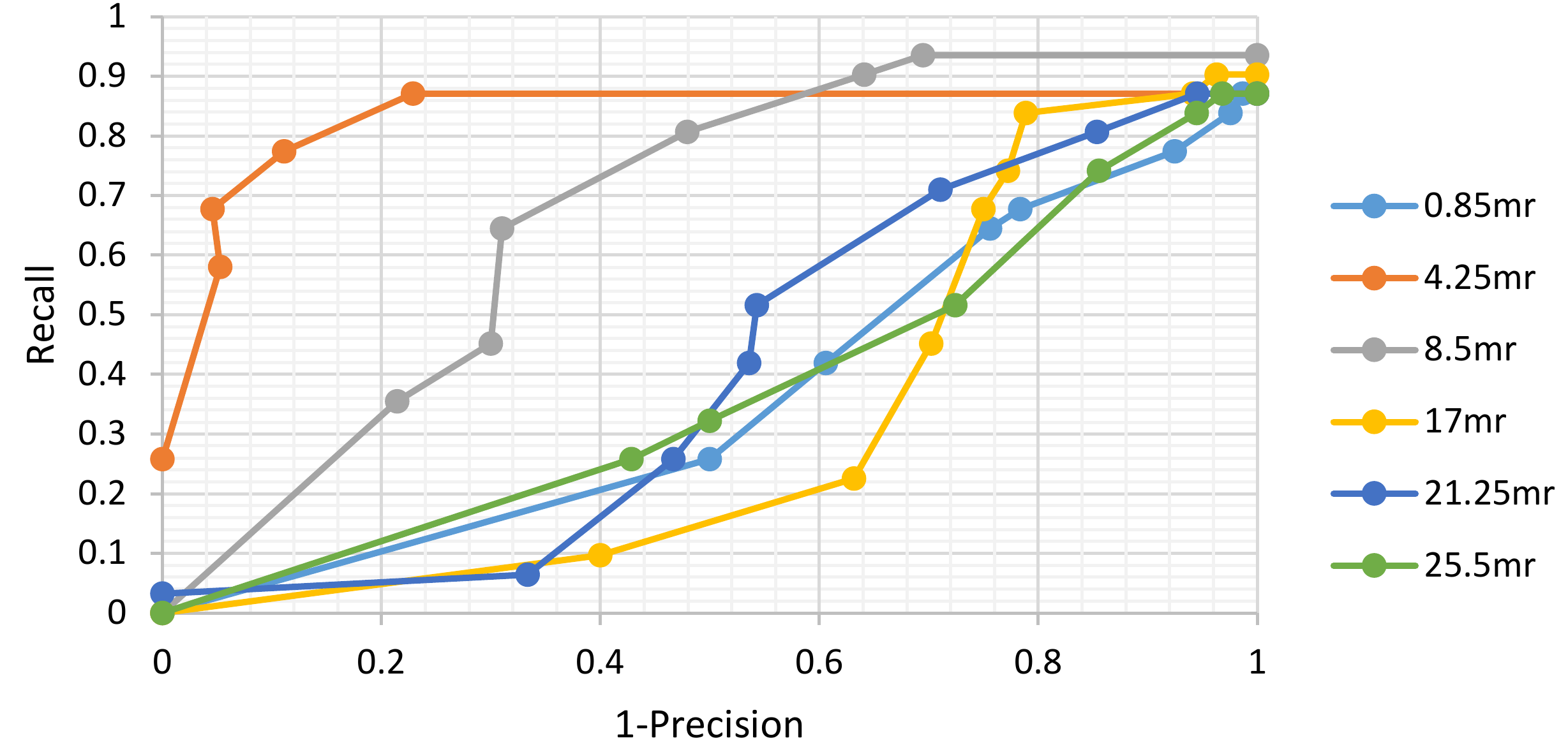}
  \includegraphics[width=4cm]{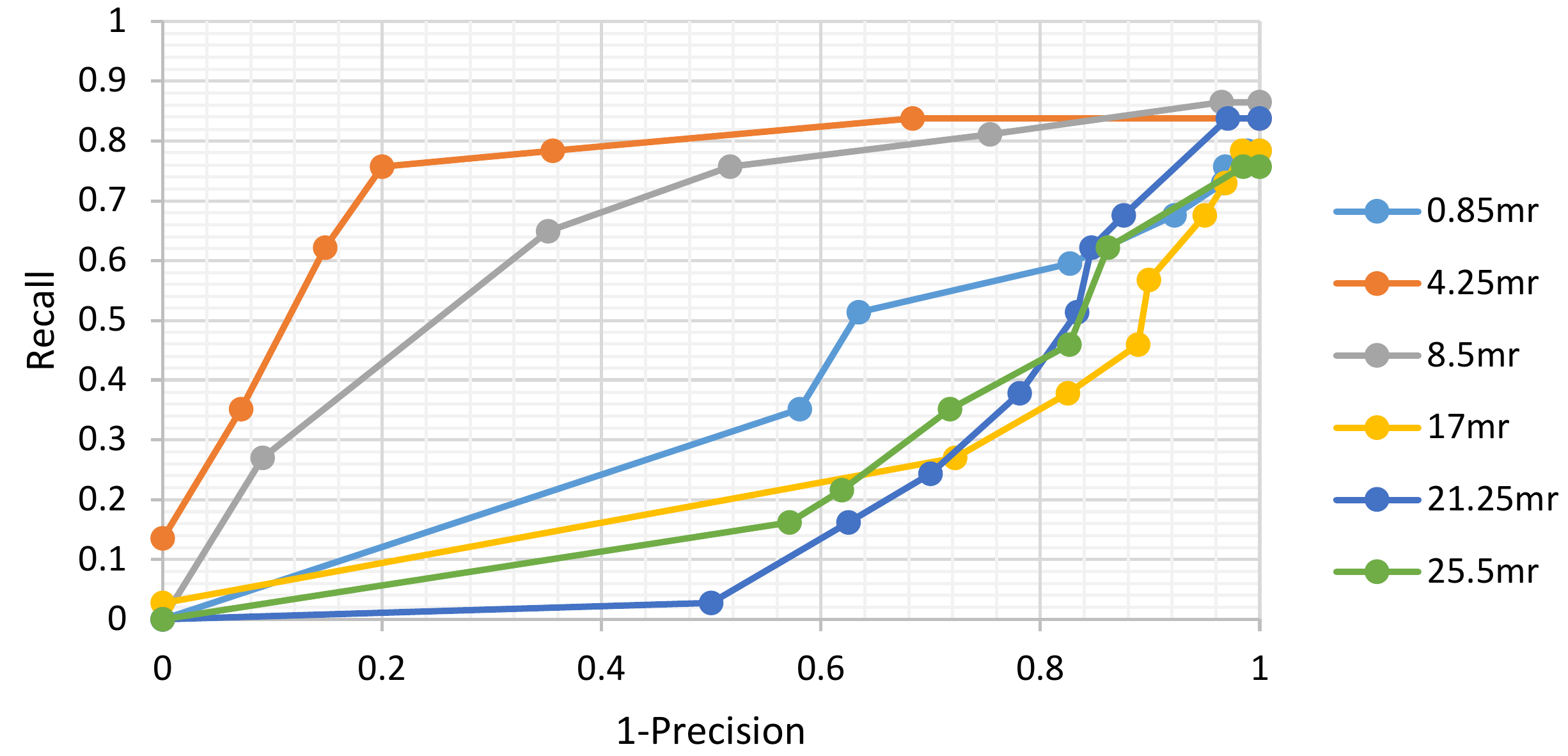}
  \caption{Feature with different support radius on the Bologna ande UWA Dataset}\label{fig:radius_uwa}
\end{figure}

\begin{figure}[tbp]
  \centering
  \includegraphics[width=.9\hsize]{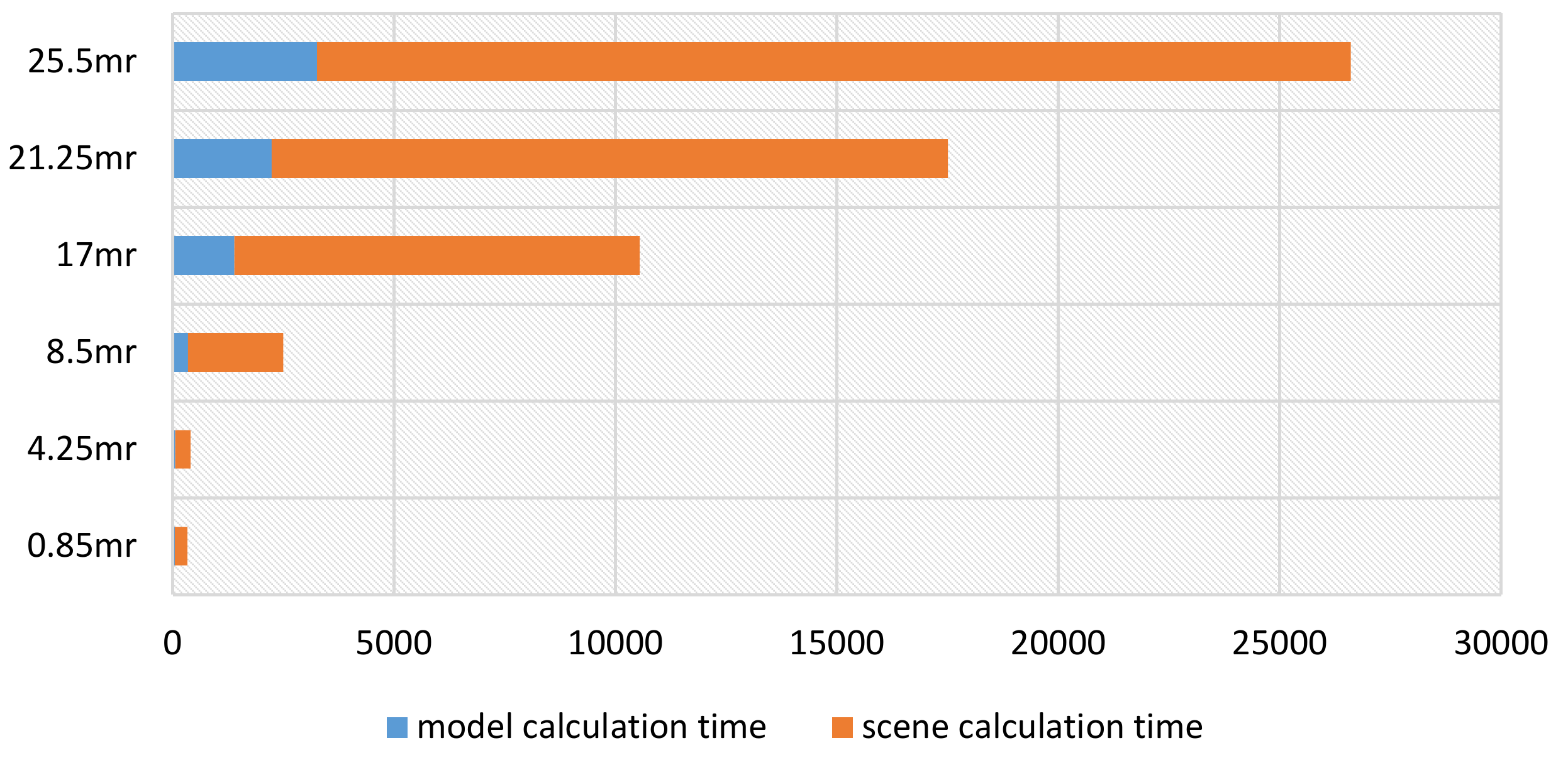}\\
  \caption{Feature calculation time with different support radius. (Unit: $ms$)}\label{fig:radius_time}
\end{figure}

We do the experiment on Bologna Dataset and UWA Dataset seperately, yielding the Recall vs 1-Precision curves in \textbf{Fig.} \ref{fig:radius_uwa}. Besides the average calculation time in models and scenes is shown in \textbf{Fig.} \ref{fig:radius_time}.  These figures clearly show two optimal support radii of 4.25mr and 8.5mr, as a tradeoff among efficiency, descriptiveness and robustness, viz. time, details and noise. For 8.5mr can obtain more higher Recall rate, finally we choose 8.5mr as our feature's support radius.

The \emph{length Gaussian weight} $\sigma_{d}$ is related to the robustness of the local feature descriptor as it determines the main point distribution information. In the experiment we evaluate the length Gaussian weight's influence to HGND.
We select the following seven different length Gaussian weights $\sigma_{d}$: 0.35, 0.5, 1, 5, 15, 45, and 500mr. 
We compare their effects by the Recall vs 1-Precision curves. In this experiment we keep the other parameters constant.

\begin{figure}[ht]
  \centering
  \includegraphics[width=4cm]{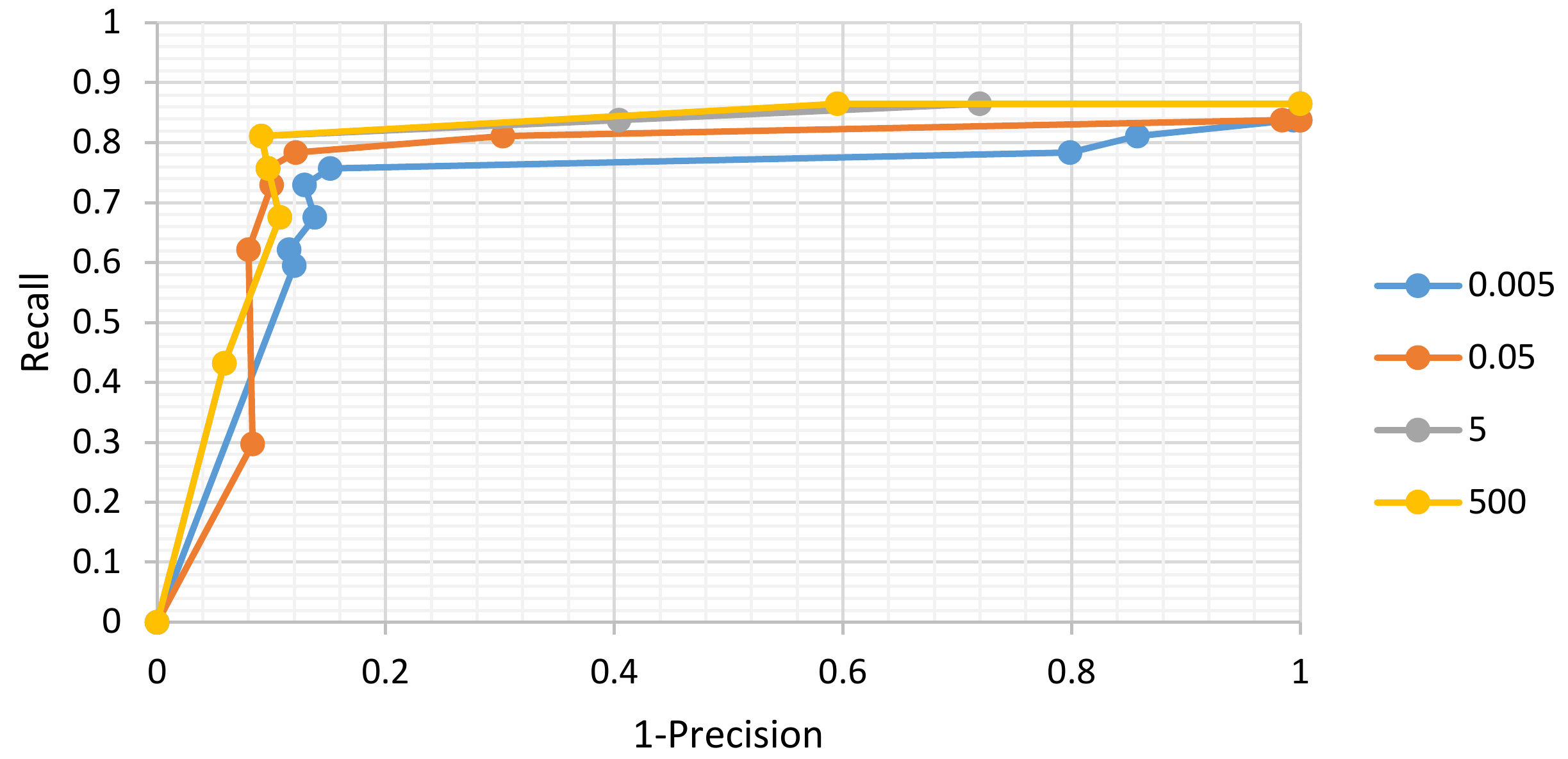}
  \includegraphics[width=4cm]{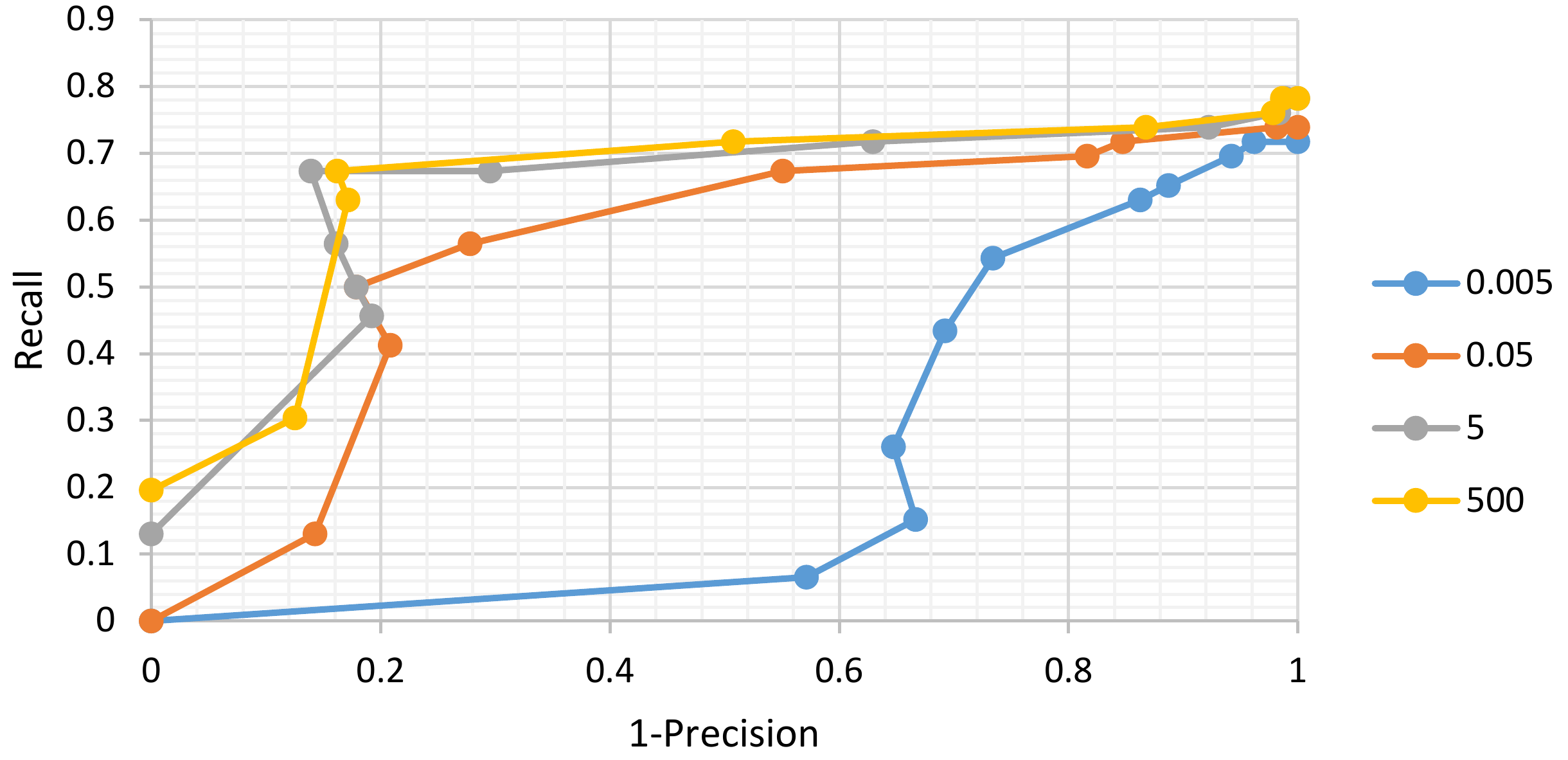}
  \caption{Feature with different direction Gaussian weight on the Bologna and UWA Dataset}\label{fig:direction_uwa}
\end{figure}

The experimental results are obtained for our ground truth perturbed by adding Gaussian noise with standard deviation $\sigma = 0.2mr$  to the surface points (See \textbf{Fig.} \ref{fig:gaussian_noise}(c)). The Recall vs 1-Precision curves for Bologna Dataset and UWA Dataset are shown in \textbf{Fig.}  \ref{fig:direction_uwa}.
From these figures it is clear that using $\sigma_{d}$ larger than $15mr$ yield best results, and the largest three $\sigma_{d} = 15mr, 45mr, 500mr$ get the same highest Recall value. But in UWA Dataset $\sigma_{d} = 500mr$ can get higher Recall value at low 1-Precision value, so we choose $500mr$ as the length Gaussian weight $\sigma_{d}$.

\begin{figure}[ht]
  \centering
  \includegraphics[width=4cm]{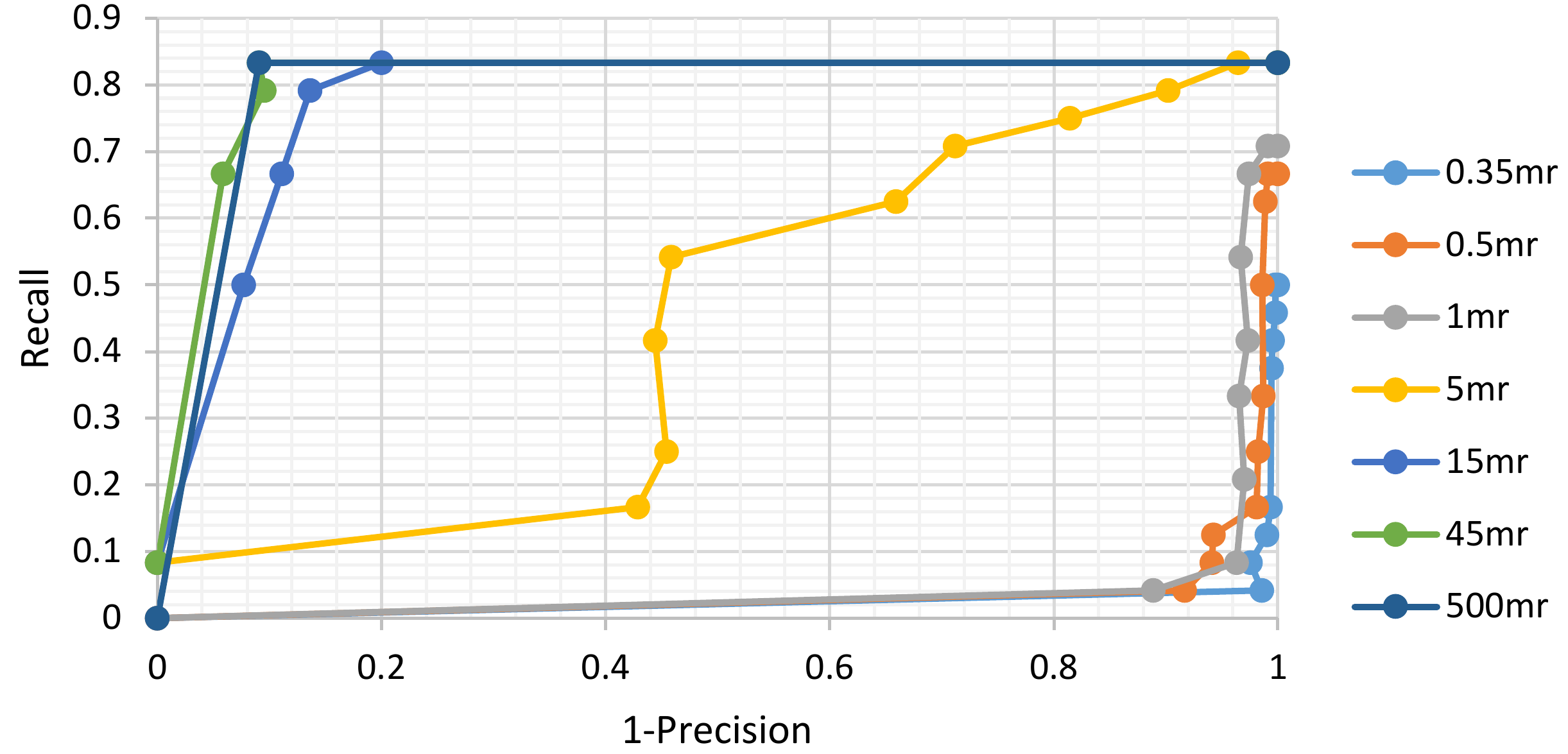}
  \includegraphics[width=4cm]{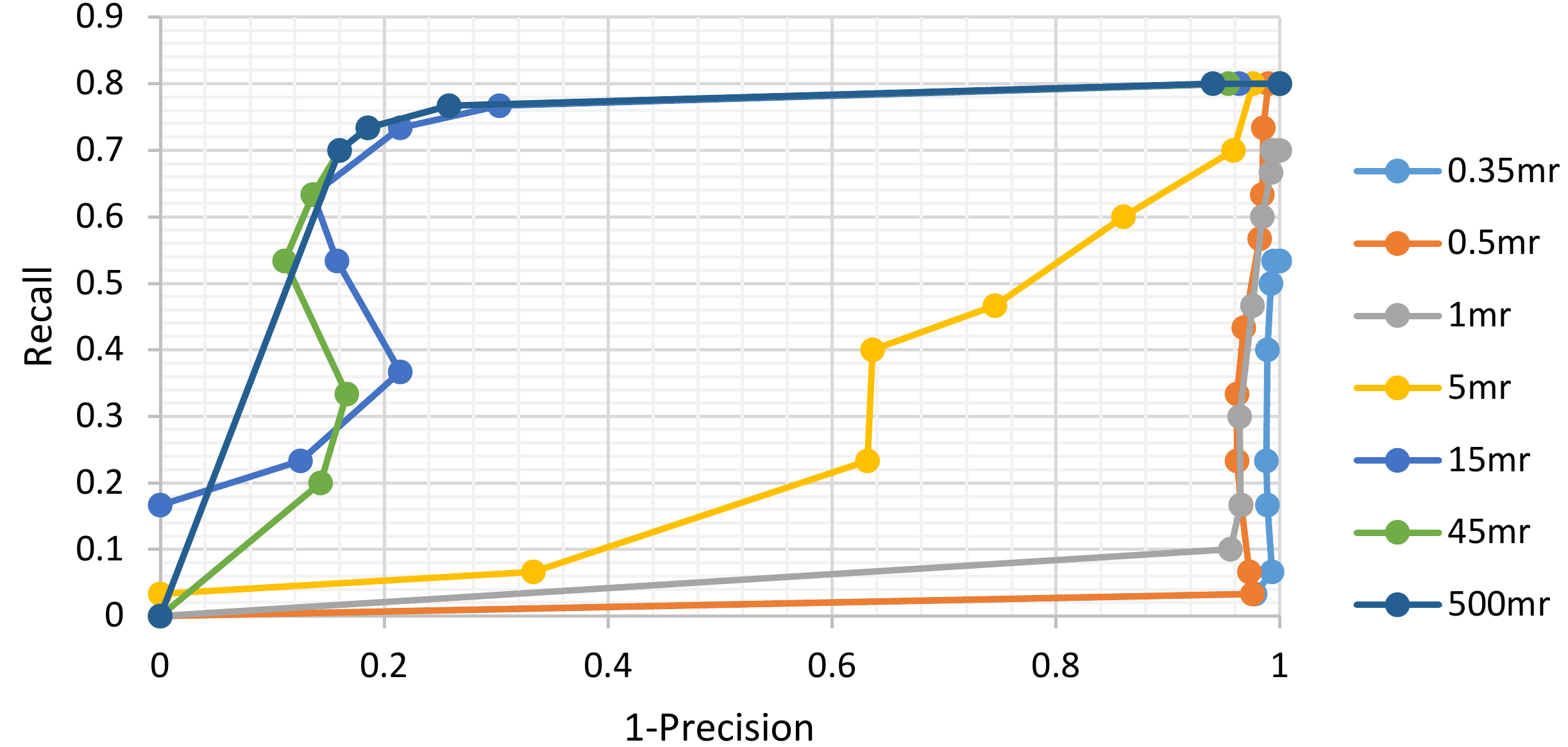}
  \caption{Feature with different length Gaussian weights on the Bologna Dataset (left) and
	UWA Dataset}\label{fig:lenght_uwa}
\end{figure}


The \emph{direction Gaussian weight} $\sigma_{\theta}$ is another important parameter for HGND's robustness to clutter, since it influence the distribution of normal. We choose $\sigma_{\theta}$ from low ($0.005$) via .05 and 5, to high ($500$) to compare their results by Recall vs 1-Precision curves, and we set other parameters constant. Other bin numbers generated worse results than those shown here. Again we used $\sigma=0.2mr$ for  Gaussian noise, yielding the Recall vs 1-Precision curves in \textbf{Fig.} \ref{fig:direction_uwa}.
These figures clearly show the more larger $\sigma_{\theta}$ implies feature more robust to clutter, when $\sigma_{\theta}$ larger than $500$, the Recall vs 1-Precision curves totally are the same. So we choose $500$ as the $\sigma_{\theta}$.

So the optimal parameters are found as support radius $r=8.5mr$, length Gaussian weight $\sigma_{d}= 500mr$, and direction Gaussian weight $\sigma_{\theta}=500$.
In Section \ref{sec:FeatureDescriptorComparison} we compare our HGND descriptor in 3D scenes  with the ROPS and SHOT descriptors.


\begin{figure}[ht]
  \centering
  \includegraphics[width=4cm]{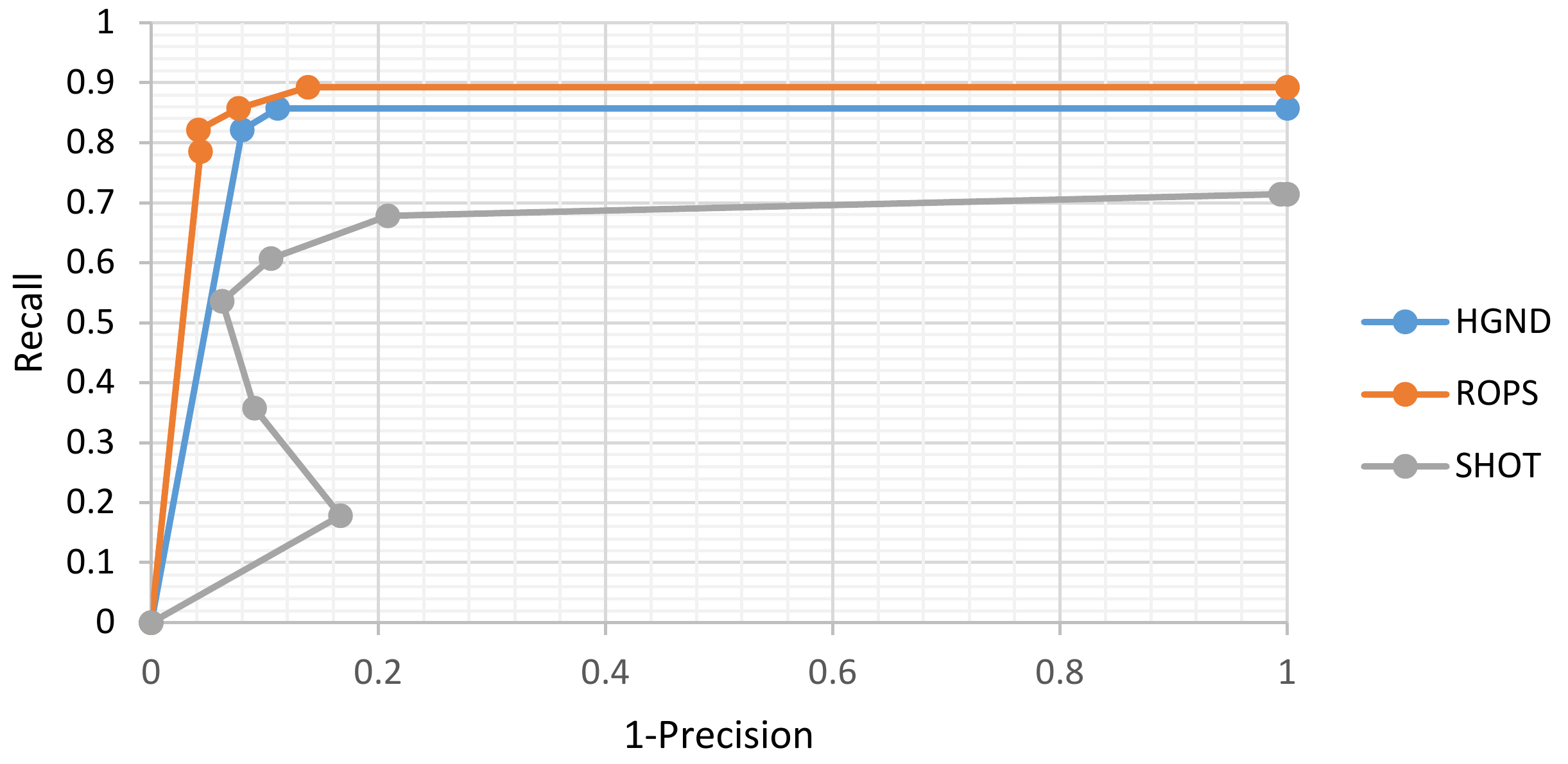}
  \includegraphics[width=4cm]{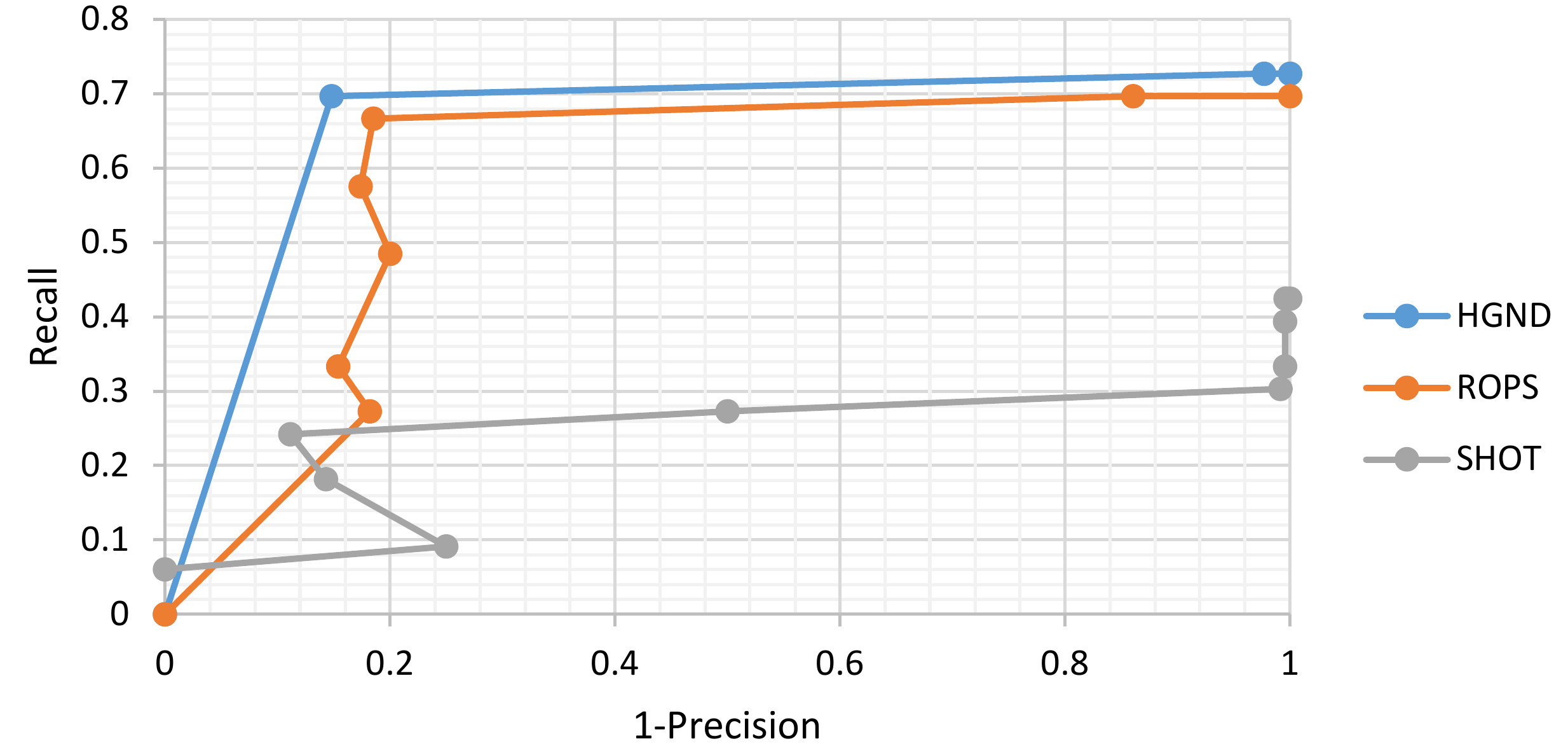}
  \caption{Feature under 0.1mr (left) and 0.3mr (right) Gaussian noise.}\label{fig:feature_0.3mr}
\end{figure}

\subsection{Feature Descriptor Comparison}
\label{sec:FeatureDescriptorComparison}

In this section we will show that the combination of our LRF and our new feature descriptor yields better overall matching results than state-of-the-art.
We compare our method against
SHOT \cite{Tobari2010Unique} and ROPS \cite{Guo2013Rotational} -- see Section \ref{sec:related_work} for details on these methods.

For the sake of preventing the influence of the selection of the keypoints onto the feature descriptor, we randomly select a set of keypoints from the scenes and the models by uniform sampling. 
We use default parameters for the descriptors, as presented in Table \ref{table:feature_parameter}. For  fairness of the feature descriptor comparison, ROPS uses the support radius mentioned in their article, SHOT uses the same support radius of HGND and the radius in bracket is using for point normal calculation in SHOT, since the SHOT descriptor need enough points to compute every point's normal, most commonly the radius for point normal calculation need to be twice as large as support radius for descriptor (We have tried the support radius mentioned in SHOT's article, but can not get better result than radius we set for them.).

\begin{table}[ht]
  \centering
  \begin{tabular}{clcc}
  \hline
  Feature &length  &neighbor radius\\
  \hline
  HGND &96  &$8.5mr$\\
  ROPS &135  &$15mr$\\
  SHOT &320  &$8.5mr$ (point normal:$17mr$)\\
  \hline
  \end{tabular}
  \caption{Parameters setting for three feature descriptors.}\label{table:feature_parameter}
\end{table}


\emph{The feature descriptors with noise:}
\textbf{Fig.} \ref{fig:feature_0.3mr} shows the results under 0.1mr and 0.3mr Gaussian noise.
For low noise ROPS performs a slightly better than HGND, but at a higher noise level, our HGND performs better than ROPS and SHOT, 
since we have used a Gaussian weight to limit the influence of Gaussian noise.

\begin{figure}[ht]
  \centering
  \includegraphics[width=8cm]{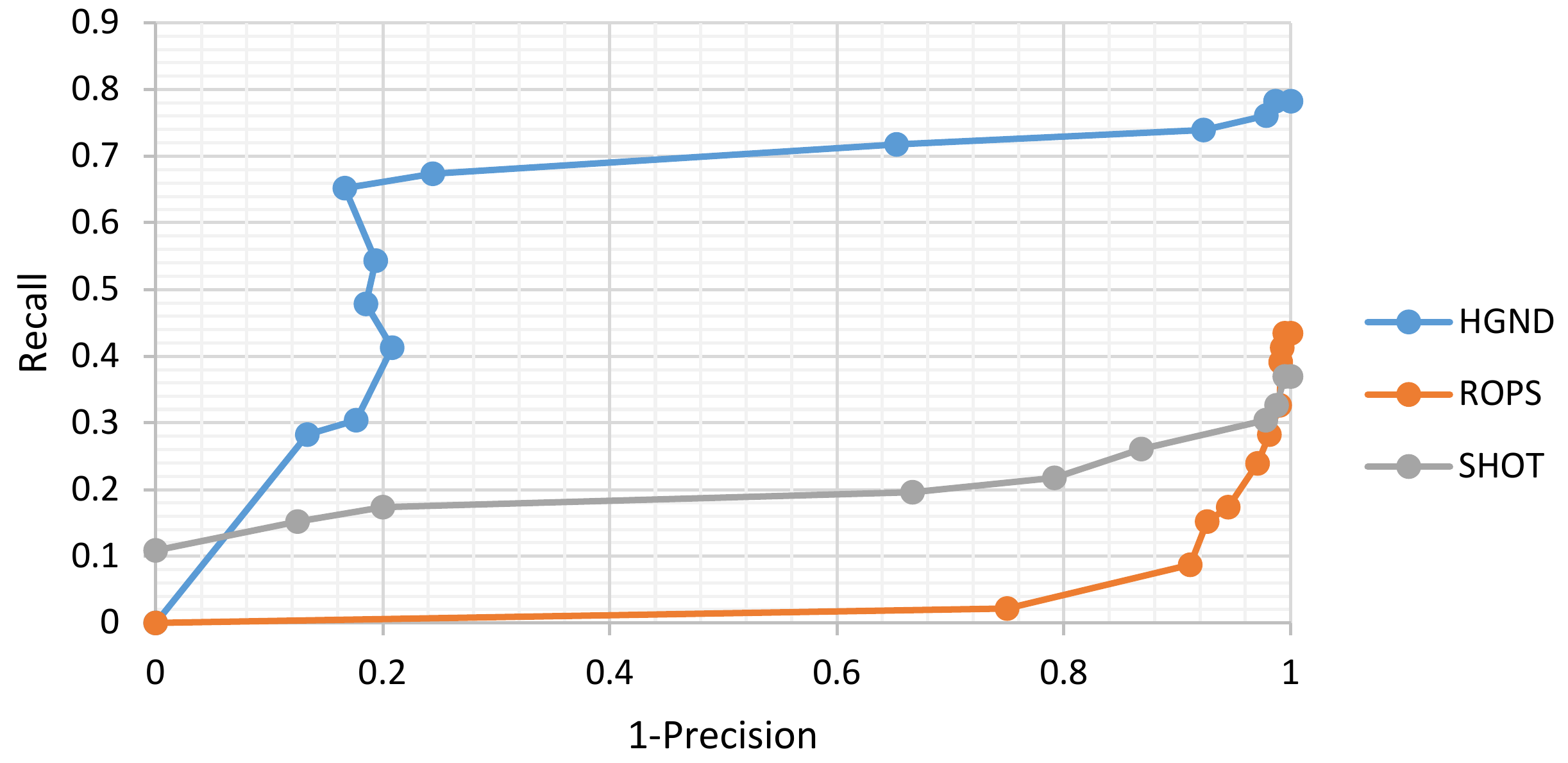}
  \includegraphics[width=8cm]{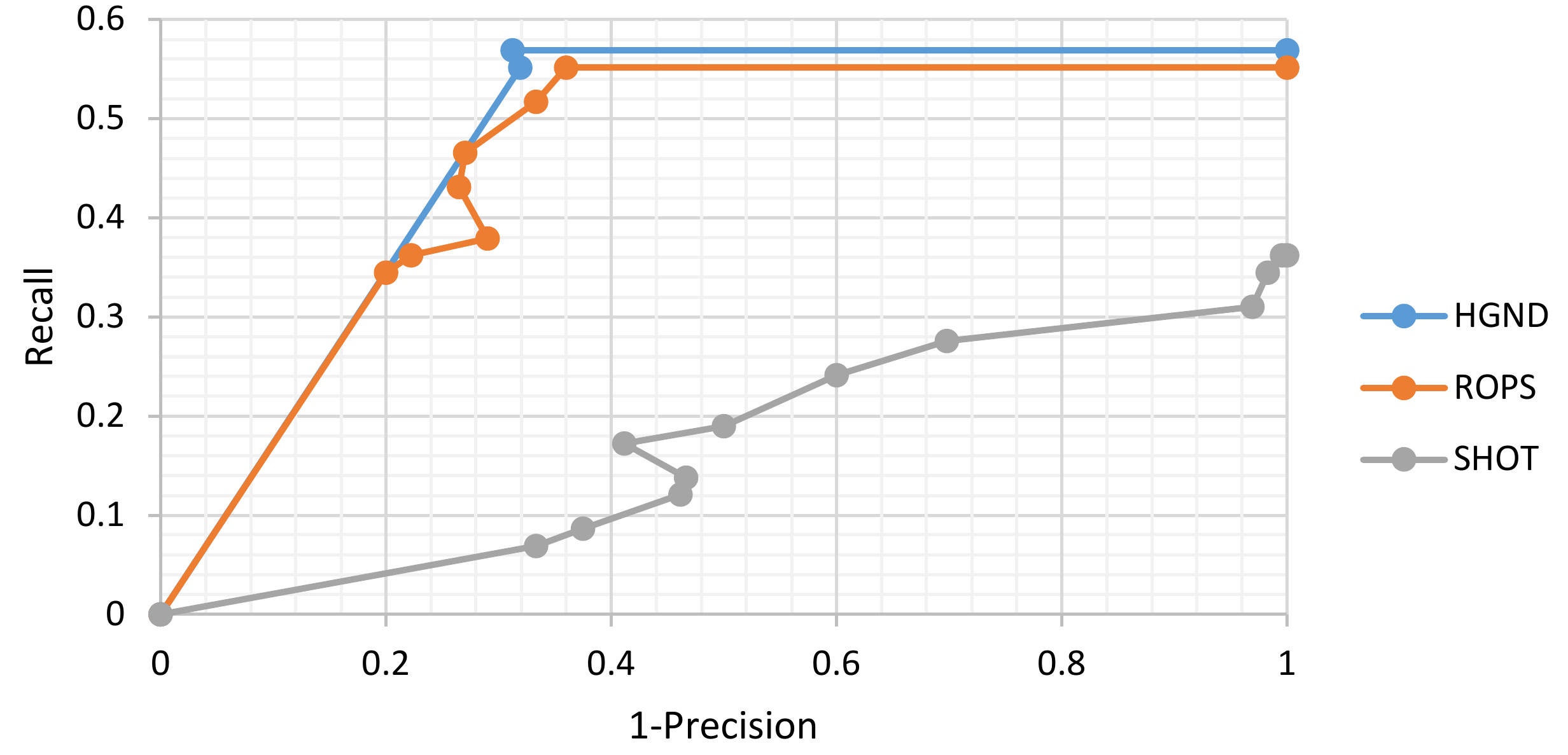}
  \caption{Feature under 1/8 sampling density (top) and under 1/2 sampling density and 0.1mr Gaussian noise (bottom).}\label{fig:feature_total}
\end{figure}

\emph{Reduced mesh resolution:}
\textbf{Fig.} \ref{fig:feature_total} (top) shows the good results of our approach for taking a mesh resolution of 1/8 compared to the original mesh resolution (See \textbf{Fig.} \ref{fig:point_density}). One sees that the low point and mesh density cause a large TP rate loss for SHOT and ROPS, especially for ROPS. 
The reason for this, is that ROPS need every point in the local surface patch to calculate feature, the low point density will result in the decreasing of feature descriptor. But our feature uses mesh center point distribution and mesh normal distribution counting obtains good result in low point density. Also the ``length" Gaussian weight and ``direction" Gaussian weight makes our descriptor invariant to the varying point and mesh density.

\emph{Both effects:}
\textbf{Fig.} \ref{fig:feature_total} (bottom) shows the results for combining the two aspects of the previous experiments: combining Gaussian noise($0.1mr$) and low point density($1/2$). Here all the methods get a big loss on the scenes both with noise and low point density, and also can observe that our descriptor outperforms the other ones gaining a TP rate up to 58$\%$.
Comparing with the previous experiments, we can find that ROPS gained very low Recall value in low point density, this is due to the fact that it relies on high point density surface making it very sensitive to low density. And SHOT is very sensitive to high Gaussian noise level since it calculates normal for every point by each point's K-neighbor points.

In these three experiments, one can see that our descriptor can gain a high recall rate near to 90\% when the noise stays at a low level. At a high noise level and a normal point density our FFIS 
can obtain an average rate about 75\%, whereas in the combined low point density scenes with a high noise level we can only get a recall rate close to 60\%.

\begin{figure}[tbp]
  \centering
  \includegraphics[width=.9\hsize]{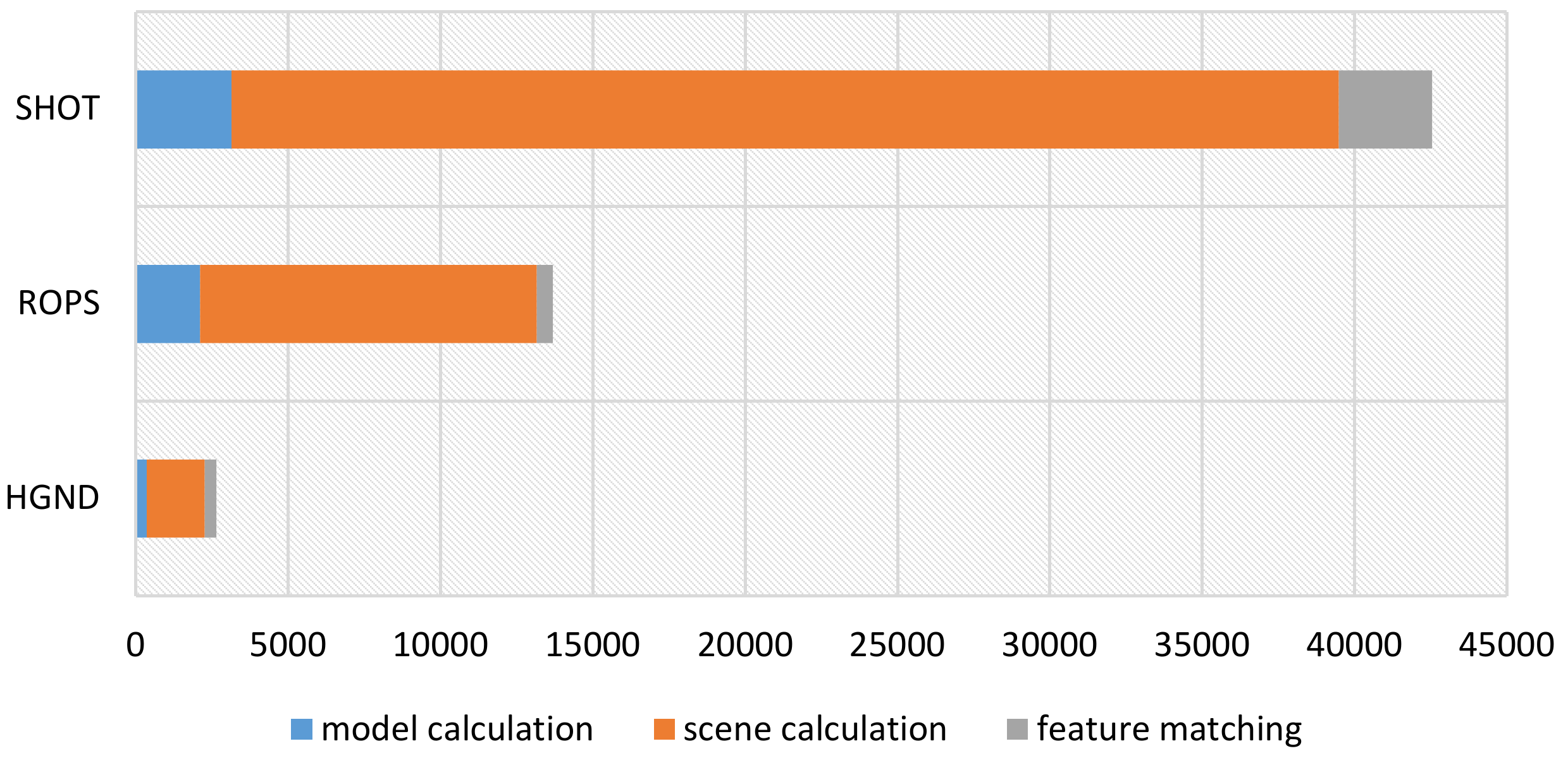}\\
  \caption{Total calculation times (in $ms$).}\label{fig:matching_time}
\end{figure}

\emph{Computation times:}
The total average calculation times for the Bologna and UWA Datasets, both for feature descriptor generation and matching, are visualized in \textbf{Fig.} \ref{fig:matching_time}.
The experiments were carried out on a computer with a Windows 10 64bit operation system, an Intel(R) Core(TM) i5-6300HQ CPU 2.30GHz processor with 12.0GB RAM. The multi-threading OPENMP (Four threads of calculation) is adopted in all the methods.
Our HGND is clearly the fastest method. It furthermore yields the best performance in noisy, cluttered scenes.

\section{Conclusion}

We considered 3D model matching where models are present in scenes but could be altered due to rotation, translation, noise, clutter, occlusion and varying point density.
To solve the problem that feature mismatching occurs we presented
a novel feature descriptor: Histograms of Gaussian Normal Distribution (HGND).

Our HGND combines geometrical information and spatial distribution information based on two Gaussian weights.
We use the transformed mesh center points and transformed mesh normals which were calculated by LRF matrix. With the point and normal transferred in LRF, making the feature descriptor easily computable and invariant to rotation and translation.
With the descriptive point distribution, normal distribution and Gaussian weights we obtain 96 dimension histograms, facilitating a better robustness to disturbances.

We performed a set of experiments on the Bologna and UWA Datasets to compare our descriptor against state-of-the-art methods under different situations with noise, clutter, occlusion and varying point density.

The results of these experiments show that HGND performs best with respect to descriptiveness and robustness to disturbances, when comparing against state-of-the-art descriptors (ROPS, SHOT). Especially under a lower noise level our HGND obtained a 90\% Recall rate. In general, our approach is able to find more true feature matchings in scenes with different disturbances, in comparison to the other approaches.

We currently focus on further research after 3D feature descriptor(eg. 3D object recognition and pose estimate). We furthermore work with data collected from a 3D scanner. 


\end{document}